\documentclass{article}

\usepackage{times}
\usepackage{graphicx} 
\usepackage{subfigure} 

\usepackage{natbib}

\usepackage{algorithm}
\usepackage{algorithmic}

\usepackage{hyperref}

\usepackage{bm}  
\usepackage{amsmath}
\usepackage{amssymb}
\usepackage{amsfonts}
\usepackage{booktabs}
\usepackage{url}

\usepackage{graphicx}

\usepackage{verbatim}

\newcommand{\mbf}[1]{{\boldsymbol{\mathbf{#1}}}}
\renewcommand{\bm}{\mbf}

\usepackage{color}

\newcommand{\Kron}{{\rm kron\_mvprod}}
\newcommand{\Reshape}{{\rm reshape}}
\newcommand{\Vect}{{\rm vec}}

\usepackage[accepted]{acm} 

\icmltitlerunning{GPatt: Fast Multidimensional Pattern Extrapolation with Gaussian Processes}

\begin{document}

\twocolumn[
\icmltitle{GPatt: Fast Multidimensional Pattern Extrapolation with Gaussian Processes}

\icmlauthor{Andrew Gordon Wilson*$^\alpha$, Elad Gilboa*$^\beta$, Arye Nehorai$^\beta$, John P.\ Cunningham$^\gamma$}{ }

\icmlkeywords{Gaussian Processes, Bayesian Nonparametrics, Kernel Learning, Feature Discovery, Extrapolation, Pattern Discovery, Gaussian Process, ICML}

\vskip 0.3in
]

\begin{abstract} 
Gaussian processes are typically used for smoothing and interpolation on small datasets.
We introduce a new Bayesian nonparametric framework -- GPatt -- enabling automatic
pattern extrapolation with Gaussian processes on large multidimensional datasets.  GPatt unifies and extends 
highly expressive kernels and fast exact inference techniques.  Without human intervention -- no hand crafting of kernel features, and no 
sophisticated initialisation procedures -- we show that GPatt can solve large scale pattern 
extrapolation, inpainting, and kernel discovery problems, including a problem with 
$383400$ training points.  We find that GPatt significantly outperforms popular alternative scalable
Gaussian process methods in speed and accuracy.  Moreover, we discover profound differences between 
each of these methods, suggesting expressive kernels, nonparametric representations, and exact 
inference are useful for modelling large scale multidimensional patterns. 
\end{abstract} 

\date{}

\section{Introduction}

``The future of the human enterprise may well depend on Big Data'', exclaimed \citet{west2013}, writing for Scientific American. 
Indeed we have quickly entered an era of \textit{big data}, focussing recent machine learning efforts on developing scalable 
models for large datasets, with notable results from deep neural architectures \citep{alex2012}.

Neural networks first became popular in the 1980s because they allowed for adaptive basis functions, as opposed to the fixed basis functions in well known 
linear models.  With adaptive basis functions, neural networks could automatically discover interesting structure in data, 
while retaining scalable learning procedures \citep{rumelhart1986learning}.  But this newfound expressive power came at the cost of interpretability and the lack of a principled 
framework for deciding upon network architecture, activation functions, learning rates, etc., all of which greatly affect performance.

Following neural networks came the kernel era of the 1990s, where infinitely many fixed basis functions were used with finite computational 
resources via the \textit{kernel trick} -- implicitly representing inner products of basis functions using a kernel.  Kernel methods are flexible,
and often more interpretable and manageable than neural network models.  For example, Gaussian processes can be used as rich prior distributions over 
functions with properties -- smoothness, periodicity, etc. -- controlled by an interpretable covariance kernel.\footnote{We use the terms \textit{covariance kernel}, \textit{covariance function}, and \textit{kernel} interchangeably.}  
Indeed Gaussian processes have had success on challenging non-linear regression and classification problems \citep{rasmussenphd96}.  

Within the machine learning community, Gaussian process research developed out of neural networks research.  \citet{neal1996} argued that since 
we can typically improve the performance of a model by accounting for additional structure in data, we ought to pursue
the limits of large models.  Accordingly, \citet{neal1996} showed that Bayesian neural networks become Bayesian nonparametric Gaussian processes with 
a \textit{neural network kernel}, as the number of hidden units approach infinity.  Thus Gaussian processes as nonparametric kernel machines are part of a 
natural progression, with the flexibility to fit any dataset, automatically calibrated complexity \citep{rasmussen06, rasmussen01}, 
easy and interpretable model specification with covariance kernels, and a principled probabilistic framework for learning kernel hyperparameters.

However, kernel machines like Gaussian processes are typically unable to scale to large modern datasets.  Methods to improve scalability usually involve 
simplifying assumptions, such as finite basis function expansions \citep{lazaro2010sparse, williams2001, le2013fastfood, rahimi07}, or sparse approximations 
using pseudo (inducing) inputs \citep{snelson2006sparse, hensman2013uai, seeger2003fast,quinonero2005unifying}.  While these methods are promising, they
simplify standard Gaussian process models, which are sometimes already too simple, particularly when a large number of training instances are available to learn
sophisticated structure in data.

Indeed popular covariance kernels used with Gaussian processes are not often expressive enough to capture rich structure in data 
and perform extrapolation, prompting \citet{mackay98} to ask whether we had ``thrown out the baby with the bathwater''.
In general, choice of kernel profoundly affects the performance of a Gaussian process -- as much as choice of architecture affects the 
performance of a neural network.  Typically, Gaussian processes are used either as flexible statistical tools, where a human manually discovers 
structure in data and then hard codes that structure into a kernel, or with the popular Gaussian (squared exponential) or Mat{\'e}rn kernels.  
In either case, Gaussian processes are used as smoothing interpolators, only able to discover limited covariance structure.  Likewise, multiple 
kernel learning \citep{gonen2011} typically involves hand crafting combinations of Gaussian kernels for specialized applications, such as modelling 
low dimensional structure in high dimensional data, and is not intended for automatic pattern discovery and extrapolation.

In this paper we propose a scalable and expressive Gaussian process framework, \textit{GPatt}, for automatic pattern discovery and 
extrapolation on large multidimensional datasets.  We begin, in Section \ref{sec: gps}, with a brief introduction to Gaussian
processes.  In Section \ref{sec: easycov} we then introduce expressive interpretable kernels, which build off the recent kernels in 
\citet{wilsonadams2013}, but are especially structured for multidimensional inputs and for the fast exact inference and learning
techniques we later introduce in Section \ref{sec: fastkernel}.  These inference techniques work by exploiting the existing structure 
in the kernels of Section \ref{sec: easycov} -- and will also work with popular alternative kernels.  These techniques relate 
to the recent inference methods of \citet{saatchi11}, but relax the full 
grid assumption made by these methods.  This exact inference and learning costs $\mathcal{O}(PN^{\frac{P+1}{P}})$ computations and 
$\mathcal{O}(PN^{\frac{2}{P}})$ storage, for $N$ datapoints and $P$ input dimensions, compared to the standard $\mathcal{O}(N^3)$ 
computations and $\mathcal{O}(N^2)$ storage associated with a Cholesky decomposition.  

In our experiments of Section \ref{sec: experiments} we combine these fast inference techniques and expressive kernels to form
GPatt.  Our experiments emphasize that, although Gaussian processes are typically only used for smoothing and 
interpolation on small datasets, Gaussian process models can in fact be developed to automatically solve a variety of practically important 
large scale pattern extrapolation problems.  GPatt is able to discover the underlying structure of an image, and extrapolate that structure
across large distances, without human intervention -- no hand crafting of kernel features, no sophisticated initialisation,
and no exposure to similar images.  We use GPatt to reconstruct large missing regions in pattern images, to restore a stained
image, to reconstruct a natural scene by removing obstructions, and to discover a sophisticated 3D ground truth kernel from movie data. GPatt
leverages a large number of training instances ($N > 10^5$) in many of these examples.

We find that GPatt significantly outperforms popular alternative Gaussian process methods on speed and accuracy stress tests.
Furthermore, we discover profound behavioural differences between each of these methods, suggesting that expressive kernels, 
nonparametric representations\footnote{For a Gaussian process to be a Bayesian nonparametric model, its kernel must be 
derived from an infinite basis function expansion.  The information capacity of such models grows with the
data \citep{ghahramani2012}.}, and exact inference -- when used together -- are useful for large scale multidimensional 
pattern extrapolation.

\section{Gaussian Processes}
\label{sec: gps}

A Gaussian process (GP) is a collection of random variables, any finite number of which have a joint Gaussian distribution.
Using a Gaussian process, we can define a distribution over functions $f(x)$,
\begin{equation}
 f(x) \sim \mathcal{GP}(m(x),k(x,x')) \,.
\end{equation}
The mean function $m(x)$ and covariance
kernel $k(x,x')$ are defined as
\begin{align}
 m(x) &= \mathbb{E}[f(x)] \,, \\
 k(x,x') &= \text{cov}(f(x),f(x')) \,,
\end{align}
where $x$ and $x'$ are any pair of inputs in $\mathbb{R}^P$.  Any collection of function values
has a joint Gaussian distribution,
\begin{align}
 [f(x_1),\dots,f(x_N)] \sim \mathcal{N}(\bm{\mu},K) \,,
\end{align}
with mean vector $\bm{\mu}_i = m(x_i)$ and $N \times N$ covariance matrix $K_{ij} = k(x_i,x_j)$.

Assuming Gaussian noise, e.g.\ observations $y(x) = f(x) + \epsilon(x)$, $\epsilon(x) = \mathcal{N}(0,\sigma^2)$,
the predictive distribution for $f(x_*)$ at a test input $x_*$, conditioned on 
$\bm{y} = (y(x_1),\dots,y(x_N))^{\top}$ at training inputs $X = (x_1,\dots,x_n)^{\top}$, is analytic and given by:
\begin{align}
 f(x_*) | x_*, X, \bm{y} & \sim \mathcal{N}(\bar{f}_*, \text{V}[f_*]) \label{eqn: predictivedist} \\   
 \bar{f}_* &= \bm{k}_*^{\top}(K+\sigma^2 I)^{-1}\bm{y} \label{eqn: predmean} \\
 \text{V}[f_*] &= k(x_*,x_*) - \bm{k}_*^{\top}(K+\sigma_n^2 I)^{-1} \bm{k}_* \,,
\end{align}
where the $N \times 1$ vector $\bm{k}_*$ has entries $(\bm{k}_*)_i = k(x_*,x_i)$.

The Gaussian process $f(x)$ can also be analytically marginalised to obtain the likelihood of 
the data, conditioned only on the hyperparameters $\bm{\theta}$ of the kernel:
\begin{equation}
 \log p(\bm{y}|\bm{\theta}) \propto -[\overbrace{\bm{y}^{\top}(K_{\bm{\theta}}+\sigma^2 I)^{-1}\bm{y}}^{\text{model fit}} + \overbrace{\log|K_{\bm{\theta}} + \sigma^2 I|}^{\text{complexity penalty}}]\,.  \label{eqn: mlikeli}
\end{equation}
This \textit{marginal likelihood} in Eq.~\eqref{eqn: mlikeli} pleasingly compartmentalises into automatically calibrated 
model fit and complexity terms \citep{rasmussen01}, and can be optimized to learn the kernel 
hyperparameters $\bm{\theta}$, or used to integrate out $\bm{\theta}$ using MCMC \citep{murray-adams-2010a}.  The problem
of model selection and learning in Gaussian processes is ``exactly the problem of finding suitable properties for the 
covariance function.  Note that this gives us a model of the data, and characteristics (such as smoothness, length-scale, etc.) 
which we can interpret.'' \citep{rasmussen06}.

The popular \textit{squared exponential} (SE) kernel has the form
\begin{equation}
 k_{\text{SE}}(x,x') = \text{exp}(-0.5 ||x-x'||^2/\ell^2) \,. \label{eqn: sekernel}
\end{equation}
GPs with SE kernels are smoothing devices, only able to learn how quickly
sample functions vary with inputs $x$, through the length-scale parameter $\ell$.

\section{Kernels for Pattern Discovery}
\label{sec: easycov}

The heart of a Gaussian process model is its kernel, which encodes 
all inductive biases -- what sorts of functions are likely under the model. Popular kernels are 
not often expressive enough for automatic pattern discovery and extrapolation.  To learn rich 
structure in data, we now present highly expressive kernels which combine with the scalable exact inference 
procedures we will introduce in Section \ref{sec: fastkernel}.

In general it is difficult to learn covariance structure from a single Gaussian process realisation,
with no assumptions.  Most popular kernels -- including the Gaussian (SE), Mat{\'e}rn, $\gamma$-exponential, 
and rational quadratic kernels \citep{rasmussen06} -- assume \textit{stationarity}, meaning that they are invariant to 
translations in the input space $x$.  In other words, any stationary kernel $k$ is a function of $\tau = x-x'$, for 
any pair of inputs $x$ and $x'$.

Bochner's theorem \citep{bochner1959lectures} shows that any stationary kernel $k(\tau)$ and its 
\textit{spectral density} $S(s)$ are Fourier duals:
\begin{align}
 k(\tau) &= \int S(s)e^{2\pi i s^{\top}\tau} ds \,, \label{eqn: kernelfourier} \\
 S(s) &= \int k(\tau)e^{-2\pi i s^{\top} \tau} d\tau \label{eqn: spectralfourier} \,.
\end{align}
Therefore if we can approximate $S(s)$ to arbitrary accuracy, then we can also approximate any stationary kernel to 
arbitrary accuracy, and we may have more intuition about spectral densities than stationary kernels.  For example, 
the Fourier transform of the popular SE kernel in Eq.~\eqref{eqn: sekernel} is a Gaussian centred at the origin.  Likewise, 
the Fourier transform of a Mat{\'e}rn kernel is a $t$ distribution centred at the origin.  These results provide the intuition 
that arbitrary additive compositions of popular kernels have limited expressive power -- equivalent to density
estimation with, e.g., scale mixtures of Gaussians centred on the origin, which is not generally a model one would use for density
estimation.  Scale-location mixtures of Gaussians, however, can approximate any distribution to arbitrary 
precision with enough components \citep{kostantinos2000}, and even with a small number of components are highly
flexible models.

Suppose that the spectral density $S(s)$ is a scale-location mixture of Gaussians,
\begin{align}
S(s) = \sum_{a=1}^{A} w_a^2 [\mathcal{N}(s; \mu_a, \sigma_a^2) + \mathcal{N}(-s; \mu_a, \sigma_a^2)]/2 \,,  \label{eqn: smixture}
\end{align}
noting that spectral densities for real data must be symmetric about $s=0$ \citep{hormander1990}, and assuming that $x$, and therefore also $s$, are in $\mathbb{R}^1$.  If we take the inverse Fourier 
transform (Eq.~\eqref{eqn: spectralfourier}) of this spectral density in Equation \eqref{eqn: smixture}, then we 
analytically obtain the corresponding spectral mixture (SM) kernel function:
\begin{align}
 k_{\text{SM}}(\tau) &= \sum_{a=1}^{A} \!w_a^2\! \exp\{-2\pi^2\tau^2 \sigma_a^2\}\cos(2\pi \tau \mu_a) \,,  \label{eqn: smkernel}
\end{align}
which was derived by \citet{wilsonadams2013}, and applied solely to simple time series examples with a small number
of datapoints.  We extend this formulation for tractability with large datasets and multidimensional inputs.

Many popular stationary kernels for multidimensional inputs decompose as a product across input dimensions.  
This decomposition helps with computational tractability -- limiting the number of hyperparameters in 
the model -- and like stationarity, provides a restriction bias that can help with learning.  
For higher dimensional inputs, $x \in \mathbb{R}^P$, we propose to leverage this useful product assumption,
inherent in many popular kernels, for a spectral mixture product (SMP) kernel
\begin{equation}
 k_{\text{SMP}}(\tau | \bm{\theta}) = \prod_{p=1}^{P} k_{\text{SM}}(\tau_p | \bm{\theta}_p) \,, \label{eqn: smproduct}
\end{equation}
where $\tau_p$ is the $p^{\text{th}}$ component of $\tau = x - x' \in \mathbb{R}^{P}$, $\bm{\theta}_p$ are the hyperparameters 
$\{\mu_a,\sigma_a^2,w_a^2\}_{a=1}^{A}$ of the $p^{\text{th}}$ spectral mixture kernel in the product of 
Eq.~\eqref{eqn: smproduct}, and $\bm{\theta} = \{\bm{\theta}_p\}_{p=1}^{P}$ are the hyperparameters of the SMP kernel. 
With enough components $A$, the SMP kernel of Eq.~\eqref{eqn: smproduct} can model any product kernel to arbitrary
precision, and is flexible even with a small number of components.  We use SMP-A as shorthand for an SMP
kernel with $A$ components in each dimension (for a total of $3PA$ kernel hyperparameters and $1$ noise hyperparameter).

\section{Fast Exact Inference}
\label{sec: fastkernel}

In this Section we present algorithms which exploit the existing structure in the SMP kernels of Section 
\ref{sec: easycov}, and many other popular kernels, for significant savings in computation and memory, but
with the same exact inference achieved with a Cholesky decomposition.  

Gaussian process inference and learning requires evaluating $(K+\sigma^2 I)^{-1}\bm{y}$ and $\log |K+\sigma^2 I|$, for 
an $N \times N$ covariance matrix $K$, a vector of $N$ datapoints $\bm{y}$, and noise variance $\sigma^2$, as in 
Equations \eqref{eqn: predmean} and \eqref{eqn: mlikeli}, respectively. For this purpose, it is standard 
practice to take the Cholesky decomposition of $(K+\sigma^2 I)$ which requires $\mathcal{O}(N^3)$ computations and $\mathcal{O}(N^2)$ 
storage, for a dataset of size $N$.  However, nearly any kernel imposes significant structure on $K$ that
is ignored by taking the Cholesky decomposition. 

For example, many kernels separate multiplicatively across $P$ input 
dimensions:
\begin{equation}
 k(x_i,x_j) = \prod_{p=1}^{P} k^{p}(x_i^p,x_j^p) \,.  \label{eqn: prodkernel}
\end{equation}
We show how to exploit this structure to perform exact inference
and hyperparameter learning in $\mathcal{O}(PN^{\frac{2}{P}})$ storage and $\mathcal{O}(PN^{\frac{P+1}{P}})$ 
operations, compared to the standard $\mathcal{O}(N^2)$ storage and $\mathcal{O}(N^3)$ operations.
We first assume the inputs $x \in \mathcal{X}$ are on a multidimensional grid (Section \ref{sec: inputgrid}), 
meaning $\mathcal{X} = \mathcal{X}_1 \times \dots \times \mathcal{X}_P \subset \mathbb{R}^P$,
and then relax this grid assumption\footnote{Note the grid does not need to be regularly spaced.} in Section \ref{sec: nogrid}.

\subsection{Inputs on a Grid}
\label{sec: inputgrid}

Many real world applications are engineered for grid structure, including 
spatial statistics, sensor arrays, image analysis, and time sampling.

Assuming a multiplicative kernel and inputs on a grid, we find\footnote{Details are in the Appendix.}
\begin{enumerate}
 \item $K$ is a Kronecker product of $P$ matrices (a \textit{Kronecker matrix}) which can undergo eigendecomposition into $QVQ^{\top}$ with only $\mathcal{O}(PN^{\frac{2}{P}})$ 
 storage and $\mathcal{O}(PN^{\frac{3}{P}})$ computations \citep{saatchi11}.\footnote{The total number of datapoints 
$N = \prod_p |\mathcal{X}_p|$, where $|\mathcal{X}_p|$ is the cardinality of
$\mathcal{X}_p$.  For clarity of presentation, we assume each $|\mathcal{X}_p|$ has equal cardinality $N^{1/P}$.}  
 \item The product of Kronecker matrices such as $K$, $Q$, or their inverses, with a vector $\bm{u}$, can be performed in $\mathcal{O}(PN^{\frac{P+1}{P}})$ operations.
\end{enumerate}

Given the eigendecomposition of $K$ as $QVQ^{\top}$, we can re-write $(K+\sigma^2 I)^{-1}\bm{y}$ and $\log |K+\sigma^2 I|$ in 
Eqs. \eqref{eqn: predmean} and \eqref{eqn: mlikeli} as
\begin{align}
(K+\sigma^2 I)^{-1}\bm{y} &= (QVQ^{\top} + \sigma^2I)^{-1}\bm{y}  \\
                          &= Q(V+\sigma^2I)^{-1}Q^{\top}\bm{y}\,,   \label{eqn: kinvy}
\end{align}
and
\begin{align}
\log |K+\sigma^2 I| = \log |QVQ^{\top} + \sigma^2 I| = \sum_{i=1}^{N} \log(\lambda_i + \sigma^2) \,,  \label{eqn: mlikelambda}
\end{align}
where $\lambda_i$ are the eigenvalues of $K$, which can be computed in $\mathcal{O}(PN^{\frac{3}{P}})$.

Thus we can evaluate the predictive distribution and marginal likelihood in Eqs.~\eqref{eqn: predictivedist} and \eqref{eqn: mlikeli} to 
perform \textit{exact} inference and hyperparameter learning,       
with $\mathcal{O}(PN^{\frac{2}{P}})$ storage and $\mathcal{O}(PN^{\frac{P+1}{P}})$ operations (assuming $P>1$).

\subsection{Missing Observations}
\label{sec: nogrid}

Assuming we have a dataset of $M$ observations which are not necessarily on a grid, we can form a complete grid using
$W$ imaginary observations, $\bm{y}_W \sim \mathcal{N}(\bm{f}_W,\epsilon^{-1}I_W)$, $\epsilon \to 0$.
The total observation vector $\bm{y} = [\bm{y}_M, \bm{y}_W]^{\top}$ has $N = M + W$ entries:
$\bm{y} = \mathcal{N}(\bm{f},{D_N})$, where
\begin{align}
\label{eqn: D_N}
 {D}_N = \left[ \begin{array}{cc}
{D}_M      &   0 \\
0   &   {\epsilon}^{-1} {I}_W      \end{array} \right],
\end{align}
and $D_M = \sigma^2 I_M$.\footnote{We sometimes use subscripts on matrices to emphasize their dimensionality:
e.g., $D_N, D_M,$ and $I_W$ are respectively $N \times N$, $M \times M$, and $W \times W$ matrices.}
The imaginary observations $\bm{y}_W$ have \textit{no corrupting effect}
on inference: the moments of the resulting predictive distribution are exactly the same as for the standard
predictive distribution in Eq.~\eqref{eqn: predictivedist}.
E.g., $(K_N + D_N)^{-1} \bm{y} = (K_M + D_M)^{-1}\bm{y}_M$.\footnote{See the Appendix for a proof.}

We use preconditioned conjugate gradients (PCG) \citep{atkinson2008} to
compute  $\left(K_N + {D}_N\right)^{-1}\bm{y}$.  We use the preconditioning matrix
${C}={D}_N^{-1/2}$ to solve ${C}^{\top}\left(K_N + {D}_N\right){C}\bm{z} = {C}^{\top}\bm{y}$. 
The preconditioning matrix $C$ speeds up convergence by ignoring the
imaginary observations $\bm{y}_W$.  Exploiting the fast multiplication of Kronecker matrices, PCG takes 
$\mathcal{O}(JPN^{\frac{P+1}{P}})$ total operations (where the number of iterations $J \ll N$) to compute 
$\left(K_N + {D}_N\right)^{-1}\bm{y}$, which allows for exact inference.

For learning (hyperparameter training) we must evaluate the marginal likelihood of Eq.~\eqref{eqn: mlikeli}.
We cannot efficiently decompose $K_M + D_M$ to compute the $\log |K_M + D_M|$ complexity penalty in the marginal likelihood,
because $K_M$ is not a Kronecker matrix, since we have an incomplete grid.  We approximate the complexity 
penalty as
\begin{align}
\log |K_M + D_M| &= \sum_{i=1}^{M} \log ({\lambda}_i^{M} + \sigma^2) \\
                 &\approx \sum_{i=1}^{M} \log(\tilde{\lambda}_i^{M} + \sigma^2)\,.
\end{align} 
We approximate the eigenvalues $\lambda_i^{M}$
of $K_M$ using the eigenvalues of $K_N$ such that $\tilde{\lambda}_i^M = \frac{M}{N} \lambda_i^N$ for $i=1,\dots,M$,
which is a particularly good approximation for large $M$ (e.g. $M>1000$) \citep{williams2001}. We emphasize that only the log determinant (complexity penalty) term in the
marginal likelihood undergoes a small approximation, and inference remains exact.

All remaining terms in the marginal likelihood of Eq.~\eqref{eqn: mlikeli} can be computed exactly and efficiently 
using PCG.  The total runtime cost of hyperparameter learning and exact inference with an incomplete grid is thus
$\mathcal{O}(PN^{\frac{P+1}{P}})$.

\section{Experiments}
\label{sec: experiments} 

In our experiments we combine the SMP kernel of Eq.~\eqref{eqn: smproduct} with the fast exact
inference and learning procedures of Section \ref{sec: fastkernel}, in a GP method we henceforth call GPatt\footnote{We write
\textit{GPatt-A} when GPatt uses an SMP-A kernel.}, 
to perform extrapolation on a variety of sophisticated patterns embedded in large datasets. 

We contrast GPatt with many alternative Gaussian process kernel methods.  In particular, we compare to the recent
sparse spectrum Gaussian process regression (SSGP) \citep{lazaro2010sparse} method, which provides fast and flexible kernel learning.
SSGP models the kernel spectrum (spectral density) as a sum of point masses, such that SSGP is a finite basis 
function model, with as many basis functions as there are spectral point masses.  SSGP is similar to the recent models of 
\citet{le2013fastfood} and \citet{rahimi07}, except it learns the locations of the point masses through marginal likelihood
optimization.  We use the SSGP implementation provided by the authors at http:\//\//www.tsc.uc3m.es\//~miguel\//downloads.php.

To further test the importance of the fast inference (Section \ref{sec: fastkernel}) used in GPatt, we compare to a GP which 
uses the SMP kernel of Section \ref{sec: easycov} but with the popular fast FITC \citep{snelson2006sparse, naish2007} inference, implemented
in GPML\footnote{http:\//\/www.gaussianprocess.org\//gpml}.  We also 
compare to Gaussian processes with the popular squared exponential (SE), rational quadratic (RQ) 
and Mat{\'e}rn (MA) (with 3 degrees of freedom) kernels, catalogued in \citet{rasmussen06}, 
respectively for smooth, multi-scale, and finitely 
differentiable functions.  Since Gaussian processes with these kernels cannot scale to the large datasets we consider, we combine
these kernels with the same fast inference techniques that we use with GPatt, to enable a 
comparison.\footnote{We also considered
the model of \citet{gpss2013}, but this model is intractable for the datasets we considered and is not structured for the fast 
inference of Section \ref{sec: fastkernel}.}

Moreover, we stress test 
each of these methods, in terms of speed and accuracy, as a function of available data and extrapolation range, number of
components, and noise.  Experiments were run on a 64bit PC, with 8GB RAM and a 2.8 GHz Intel i7 processor.

In all experiments we assume Gaussian noise, so that we can express the likelihood of the data $p(\bm{y}|\bm{\theta})$ solely as a function of kernel hyperparameters $\bm{\theta}$.
To learn $\bm{\theta}$ we optimize the marginal likelihood using BFGS.  We use a simple initialisation scheme:
any frequencies $\{\mu_a\}$ are drawn from a uniform distribution from 0 to the Nyquist frequency (1/2 the sampling rate), 
length-scales $\{1/\sigma_a\}$ from a truncated Gaussian distribution, with mean proportional to the range of the data, and weights 
$\{w_a\}$ are initialised as the empirical standard deviation of the data divided by the number of components used in the model.  
In general, we find GPatt is robust to initialisation.  

This range of tests allows us to separately understand the effects
of the SMP kernel and proposed inference methods of Section \ref{sec: fastkernel}; we will show that both are required for good performance.

\subsection{Extrapolating a Metal Tread Plate Pattern}
\label{sec: metalgrid}

\begin{figure}
\centering%
\includegraphics[scale=.9]{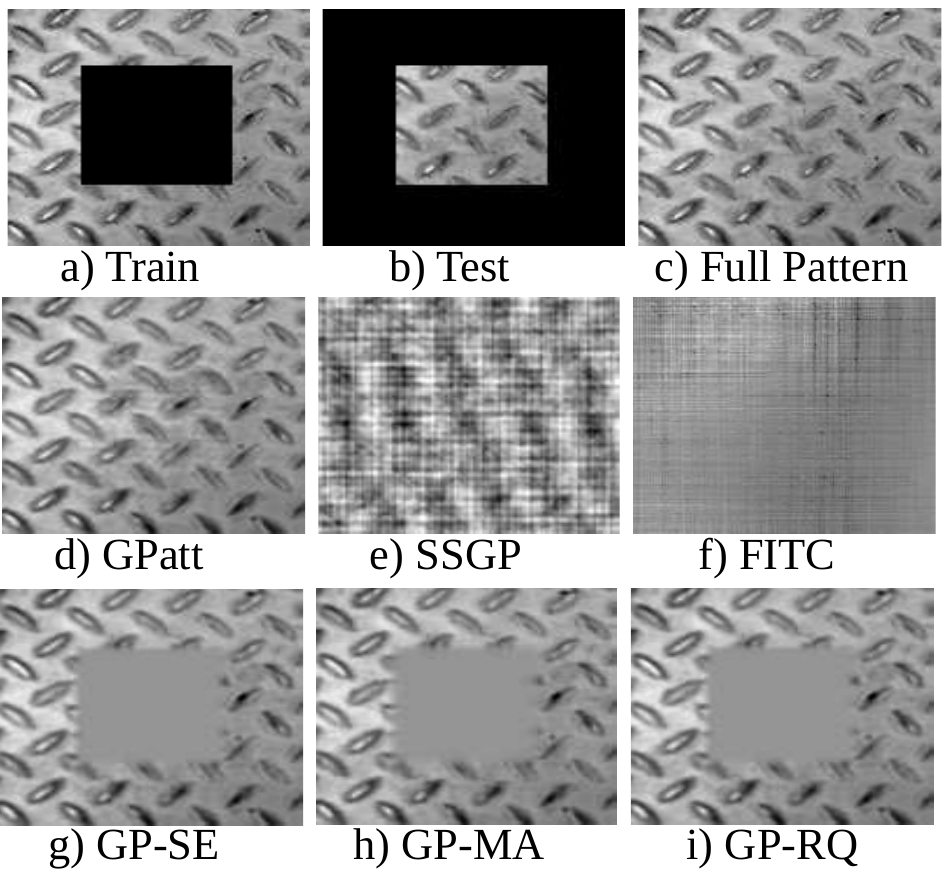}
\caption{Extrapolation on a Metal Tread Plate Pattern.  Missing data are shown in black.  a) Training region (12675 points), b) Testing region (4225 points),
c) Full tread plate pattern, d) GPatt-30, 
e) SSGP with 500 basis functions, f) FITC with 500 pseudo inputs, and the SMP-30 kernel, and GPs with the fast exact inference in 
Section \ref{sec: inputgrid}, and g) squared exponential (SE),
h) Mat{\'e}rn (MA), and i) rational quadratic (RQ) kernels.}
\label{fig: metallozenge}
\end{figure}

We extrapolate the missing region, shown in Figure \ref{fig: metallozenge}a, on a real metal tread plate texture.  There are 
12675 training instances (Figure \ref{fig: metallozenge}a), and 4225 test instances (Figure \ref{fig: metallozenge}b).  The inputs are 
pixel locations $x \in \mathbb{R}^2$ ($P=2$), and the outputs are pixel intensities. The full pattern is shown in Figure 
\ref{fig: metallozenge}c.  This texture contains shadows and subtle irregularities, no two identical diagonal markings, and patterns that have correlations 
across both input dimensions.

To reconstruct the missing region, as well as the training region, we use GPatt with $30$ components for the SMP kernel of 
Eq.~\eqref{eqn: smproduct} in each dimension (GPatt-30).  
The GPatt reconstruction shown in Figure \ref{fig: metallozenge}d 
is as plausible as the true full pattern shown in Figure \ref{fig: metallozenge}c, and largely automatic.  Without human 
intervention -- no hand crafting of kernel features to suit this image, no sophisticated initialisation, 
and no exposure to similar images -- GPatt has discovered the underlying structure of this image and 
extrapolated that structure across a large missing region, even though the structure of this pattern is not independent 
across the two spatial input dimensions.  Indeed the separability of the SMP kernel represents only a soft prior assumption, and 
does not rule out posterior correlations between input dimensions.

The reconstruction in Figure \ref{fig: metallozenge}e was produced with SSGP, using 500 basis functions.  In principle SSGP can 
model any spectral density (and thus any stationary kernel) with infinitely many components (basis functions).  However, since these components are point masses
(in frequency space), each component has highly limited expressive power.  Moreover, with many components SSGP experiences practical difficulties regarding 
initialisation, over-fitting, and computation time (scaling quadratically with the number of basis functions).  Although SSGP does discover some interesting 
structure (a diagonal pattern), and has equal training and test performance, it is unable to capture enough information for a convincing reconstruction,
and we did not find that more basis functions improved performance.  
Likewise, FITC with an SMP-30 kernel and 500 pseudo-inputs cannot capture the necessary information to interpolate or extrapolate.  
We note FITC and SSGP-500 respectively took 2 days and 1 hour to run on this example, compared to GPatt which took under 5 minutes.

GPs with SE, MA, and RQ kernels are all truly Bayesian 
nonparametric models -- these kernels are derived from infinite basis function expansions.  Therefore, as seen in Figure \ref{fig: metallozenge} g), h), i), these methods are completely able to capture 
the information in the training region; however, these kernels do not have the proper structure to reasonably extrapolate across
the missing region -- they simply act as smoothing filters.  We note that this comparison is only possible because these GPs are using the 
fast exact inference techniques in Section \ref{sec: fastkernel}.  

Overall, these results indicate that both expressive nonparametric kernels, 
such as the SMP kernel, and the specific fast inference in Section \ref{sec: fastkernel}, are needed to be able to extrapolate patterns in 
these images.

\begin{figure}
\centering%
\includegraphics[scale=.57]{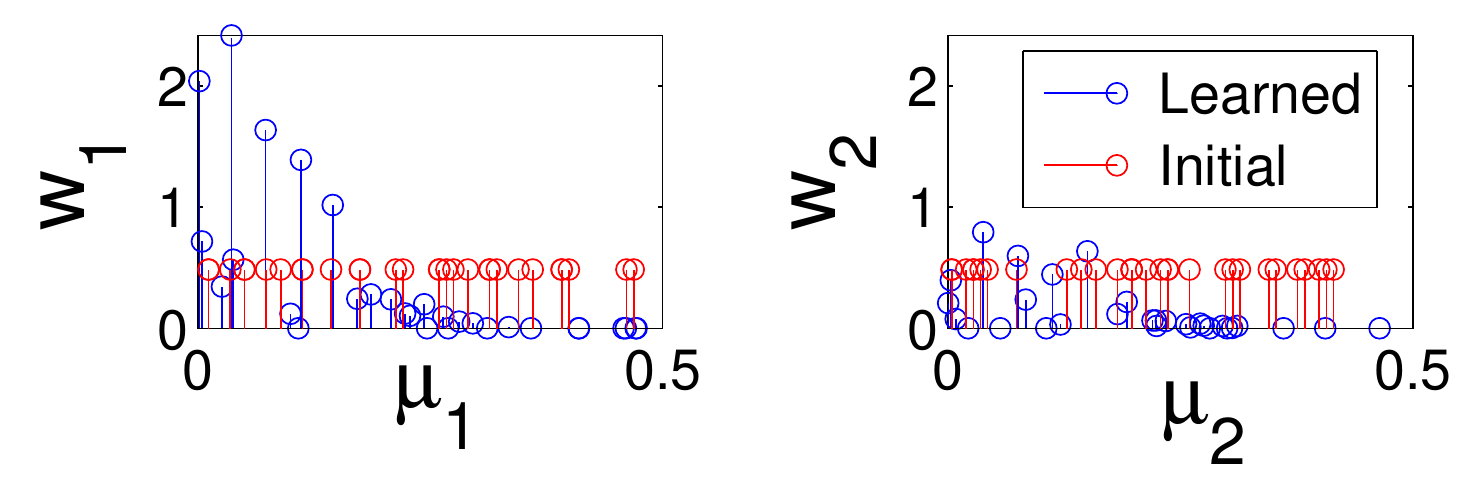}
\caption{Automatic Model Selection in GPatt.
Initial and learned weight and frequency parameters of GPatt-30,
for each input dimension (a dimension is represented in each panel), on the metal tread plate pattern 
of Figure \ref{fig: metallozenge}.  GPatt-30 is overspecified for this pattern.  
During training, weights of extraneous components automatically shrink to zero, which helps indicate
whether the model is overspecified, and helps mitigate the effects of model overspecification.
Of the 30 initial components in 
each dimension, 15 are near zero after training.}  

\label{fig: w1w2}
\end{figure}

We note that the SMP-30 kernel used with GPatt has more components than needed for this problem.
However, as shown in Figure \ref{fig: w1w2}, if the 
model is overspecified, the complexity penalty in the marginal likelihood shrinks the weights ($\{w_a\}$ 
in Eq.~\eqref{eqn: smkernel}) of extraneous components, as a proxy for model selection -- an effect 
similar to \textit{automatic relevance determination} \citep{mackay1994}.  As per Eq.~\eqref{eqn: mlikelambda}, this complexity penalty 
can be written as a sum of eigenvalues of a covariance matrix $K$.  Components which do not significantly 
contribute to model fit will therefore be automatically pruned, as shrinking the weights decreases the 
eigenvalues of $K$ and thus minimizes the complexity penalty.  This weight shrinking helps mitigate
the effects of model overspecification and helps indicate whether the model is overspecified.  In the following
stress tests we find that GPatt scales efficiently with the number of components in its SMP kernel.

\subsection{Stress Tests}
\label{sec: stresstest}

We stress test GPatt and alternative methods in terms of speed and accuracy, with varying datasizes, extrapolation ranges, 
basis functions, pseudo inputs, and components.  We assess accuracy using standardised mean square error (SMSE) and mean standardized 
log loss (MSLL) (a scaled negative log likelihood), as defined in \citet{rasmussen06} on page 23. Using the empirical mean and variance to fit the
data would give an SMSE and MSLL of 1 and 0 respectively.  Smaller SMSE and more negative MSLL values correspond to better fits of the data.

\begin{figure}
\centering%
\includegraphics[scale=.54]{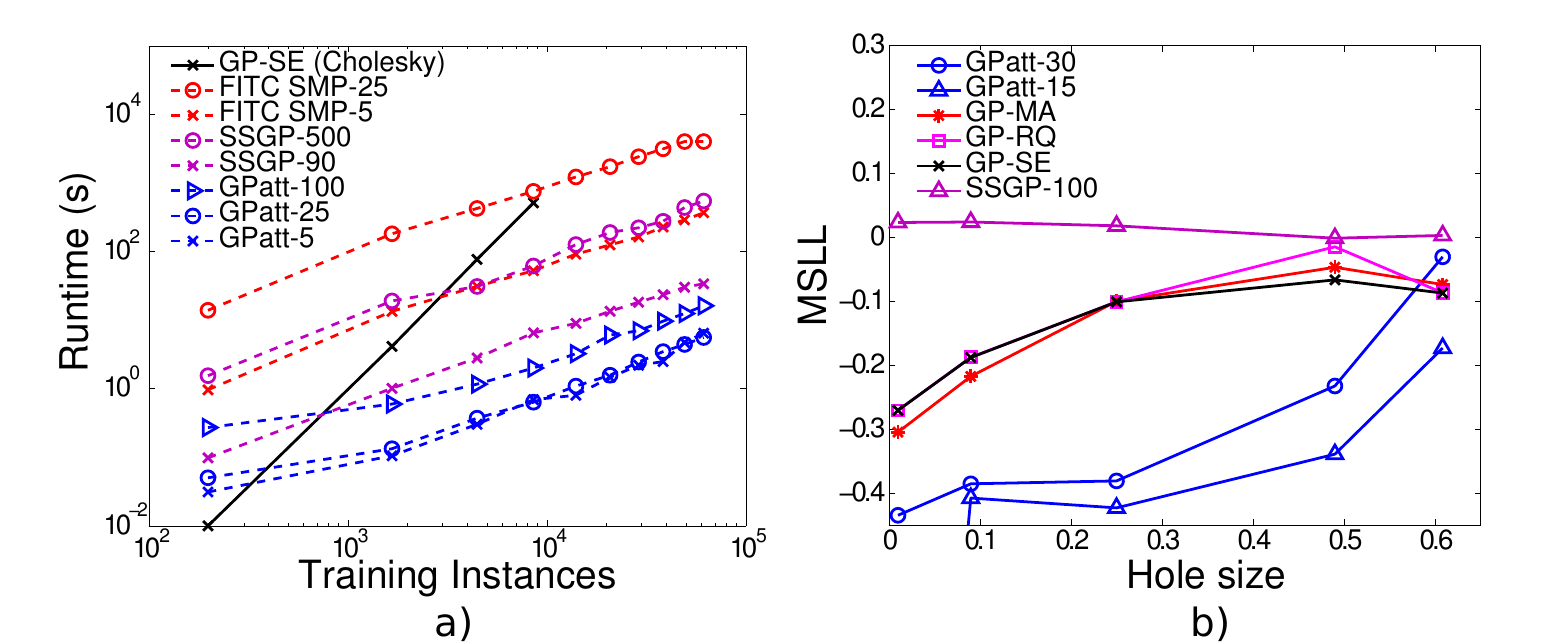}
\caption{Stress Tests.  a) \textbf{Runtime Stress Test}.  We show the runtimes in seconds, as a function of training instances, for evaluating the log marginal likelihood, 
and any relevant derivatives, for a standard GP with SE kernel (as implemented in GPML), FITC with 500 pseudo-inputs and SMP-25 and SMP-5 kernels, 
SSGP with 90 and 500 basis functions, and GPatt-100, GPatt-25, and GPatt-5.  Runtimes are for a 64bit PC, with 8GB RAM and a 2.8 GHz Intel i7 processor, on the cone pattern ($P=2$), shown in the Appendix.  The ratio of training inputs to the sum of imaginary and training inputs for GPatt 
(Section \ref{sec: nogrid}) is $0.4$ and $0.6$ for the smallest two training sizes, and $0.7$ for all other training sets.  
b) \textbf{Accuracy Stress Test}.  MSLL as a function of holesize on the metal pattern of Figure \ref{fig: metallozenge}.  The values on the horizontal axis 
represent the fraction of missing (testing) data from the full pattern (for comparison Fig \ref{fig: metallozenge}a has 25\% missing data).  
We compare GPatt-30 and GPatt-15 with GPs with SE, MA, and RQ kernels (and the 
inference of Section \ref{sec: fastkernel}), and SSGP with $100$ basis functions. The MSLL for GPatt-15 at a holesize of $0.01$ is $-1.5886$.}
\label{fig: stresstest}
\end{figure}

The runtime stress test in Figure \ref{fig: stresstest}a shows that the number of components used in GPatt 
does not significantly affect runtime, and that GPatt is much faster than FITC (using 500 pseudo inputs) and SSGP (using 90 or 500
basis functions), even with $100$ components ($601$ kernel hyperparameters).  The slope of each curve roughly indicates the asymptotic
scaling of each method.  In this experiment, the standard GP (with SE kernel) has a slope of $2.9$, which is close to the cubic scaling 
we expect.  All other curves have a slope of $1 \pm 0.1$, indicating linear scaling with the number of training instances.  However,
FITC and SSGP are used here with a \textit{fixed} number of pseudo inputs and basis functions.  More pseudo inputs and basis functions
should be used when there are more training instances -- and these methods scale quadratically with pseudo inputs and basis functions for a 
fixed number of training instances.  GPatt, on the other hand, can scale linearly in runtime as a function of training size, without any 
deterioration in performance.  Furthermore, the big gaps between each curve -- the fixed 1-2 orders of magnitude GPatt outperforms
alternatives -- is as practically important as asymptotic scaling.

The accuracy stress test in Figure \ref{fig: stresstest}b shows extrapolation (MSLL) performance on the metal tread plate pattern of Figure \ref{fig: metallozenge}c with varying holesizes,
running from 0\% to 60\% missing data for testing (for comparison the hole shown in Figure \ref{fig: metallozenge}a is for 25\% missing data).  GPs with SE, RQ, and MA kernels
(and the fast inference of Section \ref{sec: fastkernel}) all steadily increase in error as a function of holesize.  Conversely, SSGP does not increase
in error as a function of holesize -- with finite basis functions SSGP cannot extract as 
much information from larger datasets as the alternatives.  GPatt performs well relative to the other methods, even with a small number of 
components.  GPatt is particularly able to exploit the extra information in additional
training instances:  only when the holesize is so large that over 60\% of the data are missing does GPatt's performance degrade to the same level as alternative 
methods.

In Table \ref{tab: predictions} we compare the test performance of GPatt with SSGP, and GPs using SE, MA, and RQ kernels, for extrapolating five
different patterns, with the same train test split as for the tread plate pattern in Figure \ref{fig: metallozenge}.  All patterns are shown in the Appendix.
GPatt consistently has the lowest standardized mean squared error (SMSE), and mean 
standardized log loss (MSLL).  Note that many of these datasets are sophisticated patterns, containing intricate details and subtleties which are not strictly periodic, such
as lighting irregularities, metal impurities, etc.  Indeed SSGP has a periodic kernel (unlike the SMP kernel which is not strictly periodic), and is capable of 
modelling multiple periodic components, but does not perform as well as GPatt on these examples.

\begin{table}
\caption{We compare the test performance of GPatt-30 with 
SSGP (using 100 basis functions), and GPs using squared exponential (SE), Mat{\'e}rn (MA), and rational quadratic 
(RQ) kernels, combined with the inference of Section \ref{sec: stresstest}, on patterns with a train test split as
in the metal treadplate pattern of Figure \ref{fig: metallozenge}.}
\scriptsize
\begin{center}
\begin{tabular}{l r r r r r }
\toprule
  &  GPatt & SSGP & SE & MA & RQ\\
\midrule
  \multicolumn{6}{l}{\includegraphics[scale=.08]{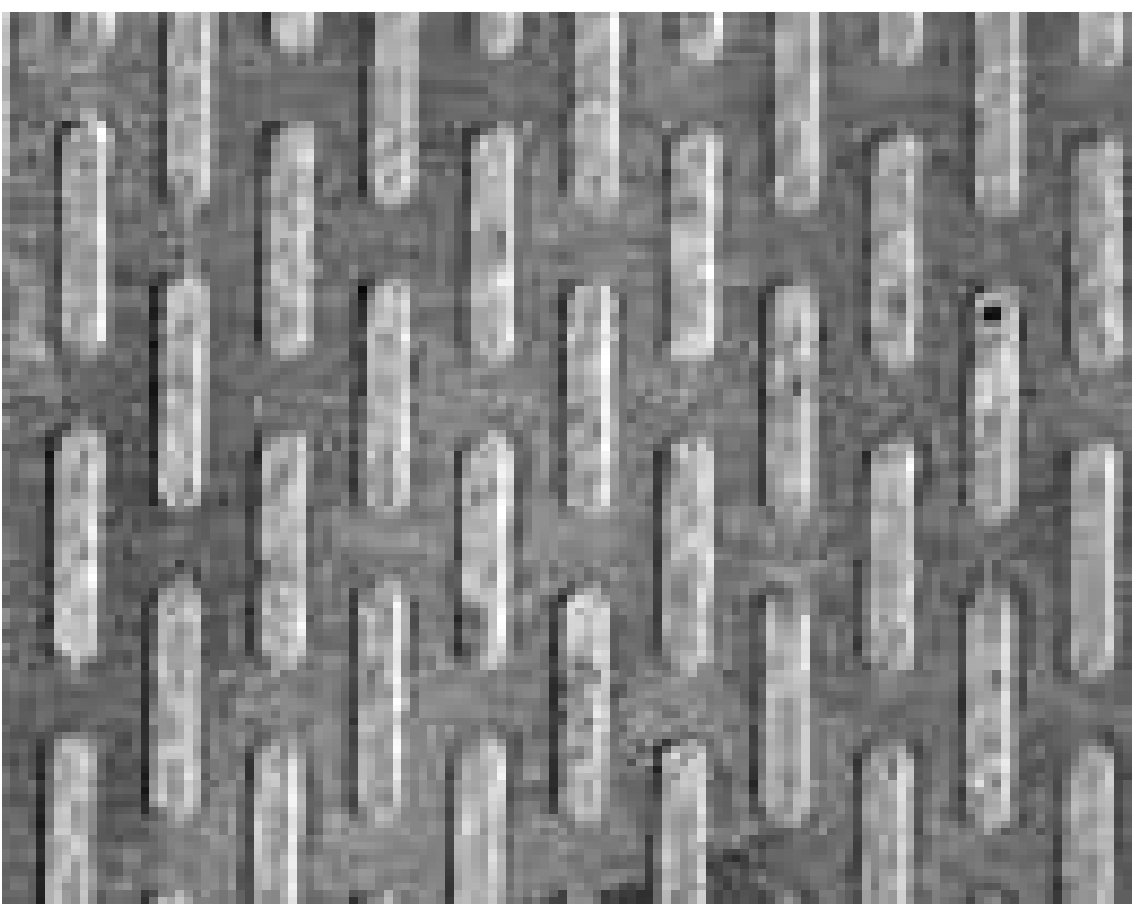} ~~{\normalsize Rubber mat}  \quad ~~~{(train = 12675, test = 4225)}}   \\
\midrule
  SMSE & $\bm{0.31}$ & $0.65$ & $ 0.97$ & $0.86$ & $ 0.89$ \\
  MSLL & $\bm{-0.57}$ & $-0.21$ & $0.14$ & $ -0.069$ & $0.039$ \\
\midrule
  \multicolumn{6}{l}{\includegraphics[scale=.08]{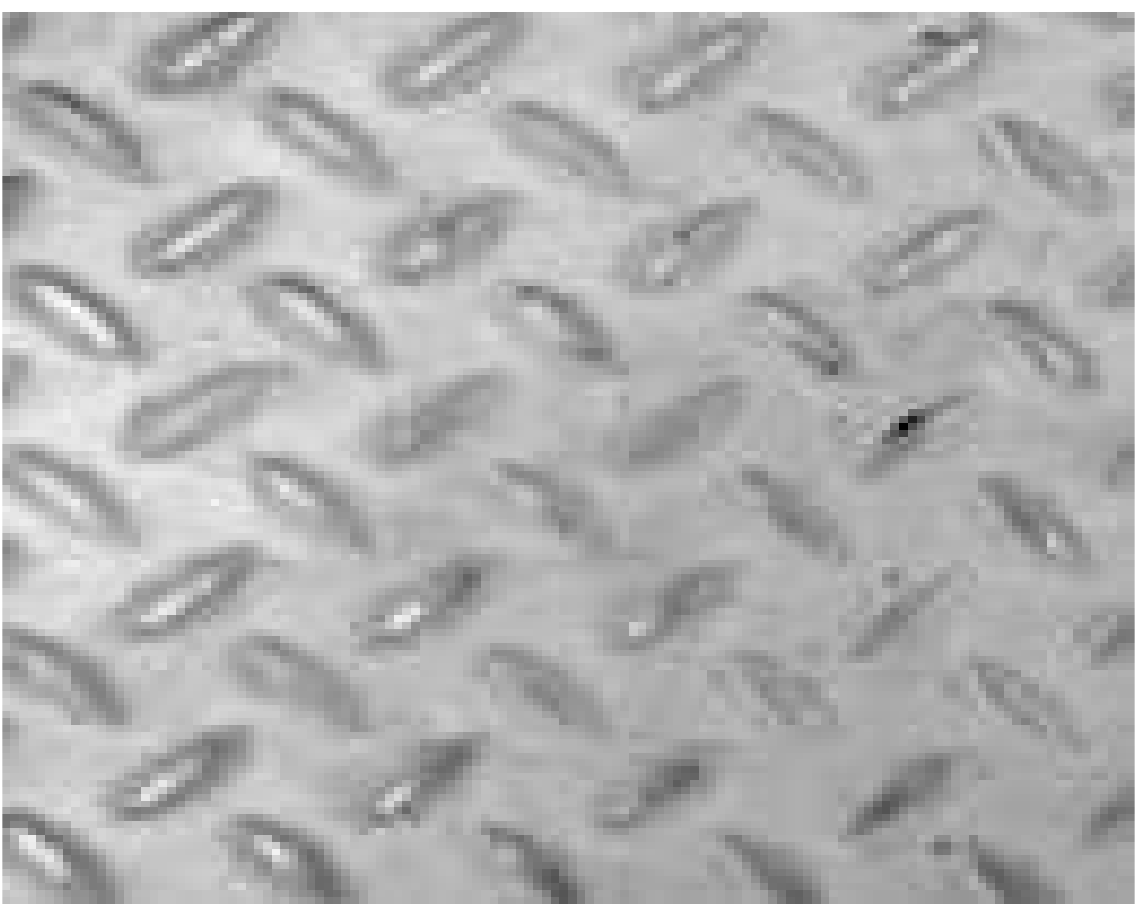} ~~{\normalsize Tread plate} \quad \text{ }~~~{(train = 12675, test = 4225)}}   \\
\midrule
  SMSE & $\bm{0.45}$ & $1.06$ & $0.895$ & $0.881$ & $0.896$ \\
  MSLL & $\bm{-0.38}$ & $0.018$ & $-0.101$ & $-0.1$ & $-0.101$ \\
\midrule
  \multicolumn{6}{l}{\includegraphics[scale=.08]{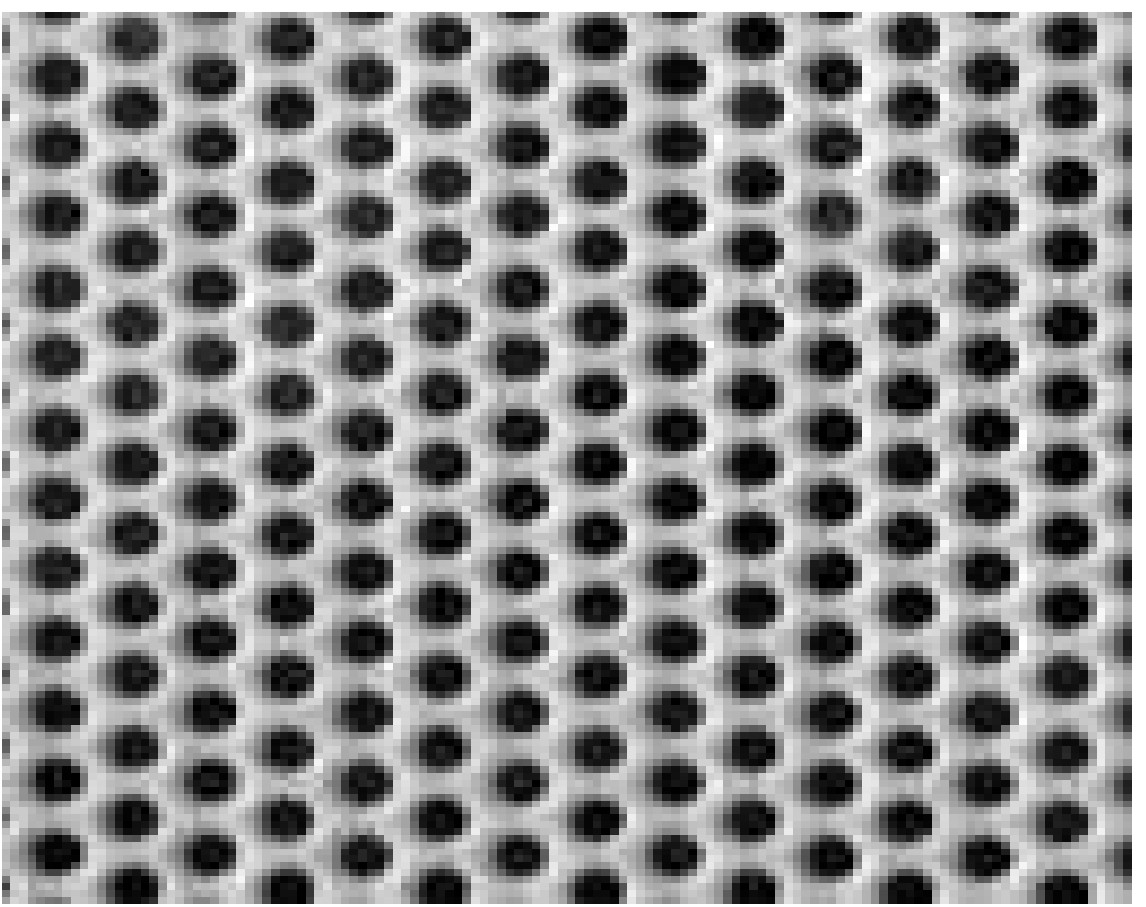} ~~ {\normalsize Pores} \qquad \qquad ~~~{(train = 12675, test = 4225)}}   \\
\midrule
  SMSE & $\bm{0.0038}$ & $1.04$ & $0.89$ & $0.88$ & $0.88$  \\
  MSLL & $\bm{-2.8}$ & $-0.024$ & $-0.021$ & $-0.024$ & $-0.048$\\
\midrule
  \multicolumn{6}{l}{\includegraphics[scale=.08]{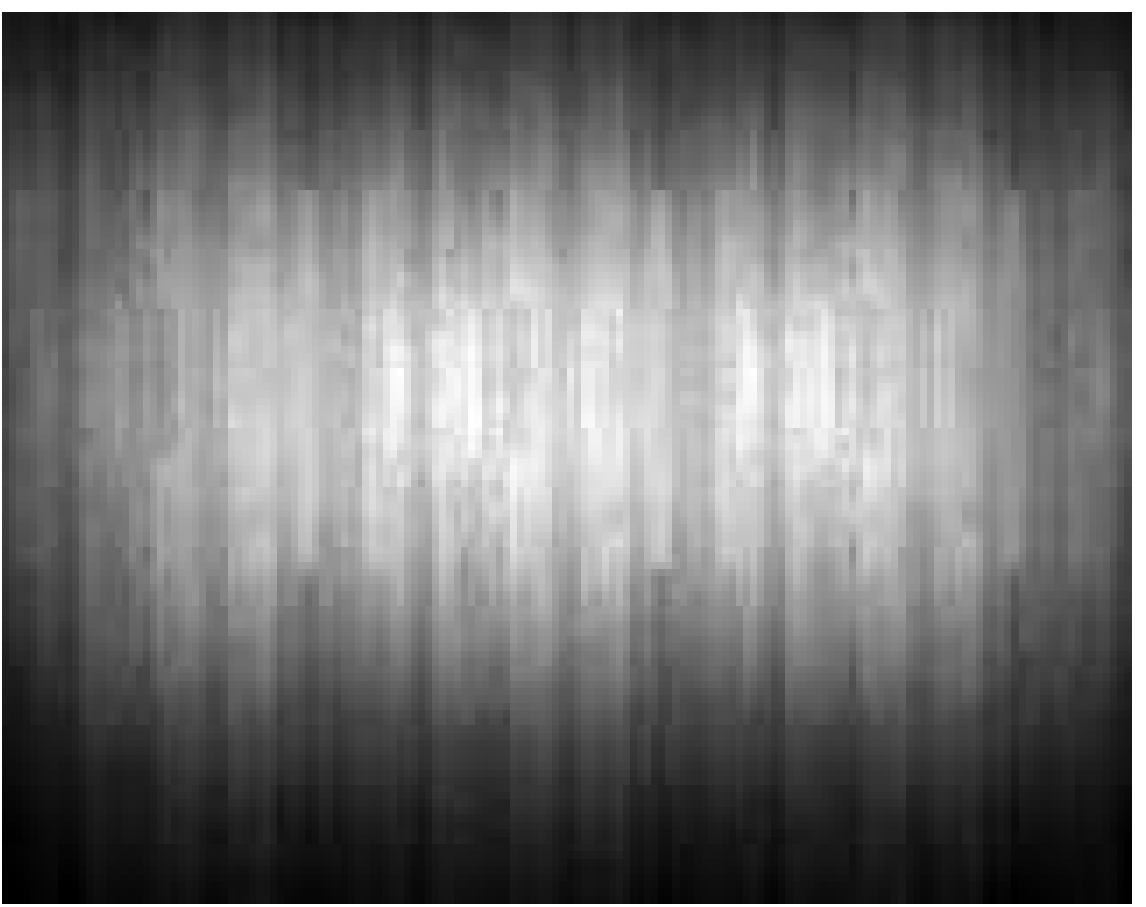} ~~ {\normalsize Wood} \qquad \qquad ~~~{(train = 14259, test = 4941)}}   \\
\midrule
  SMSE & $\bm{0.015}$ & $0.19$ & $0.64$ & $0.43$ & $0.77$  \\
  MSLL & $\bm{-1.4}$ & $-0.80$ & $1.6$ & $1.6$ & $0.77$\\
\midrule
  \multicolumn{6}{l}{\includegraphics[scale=.08]{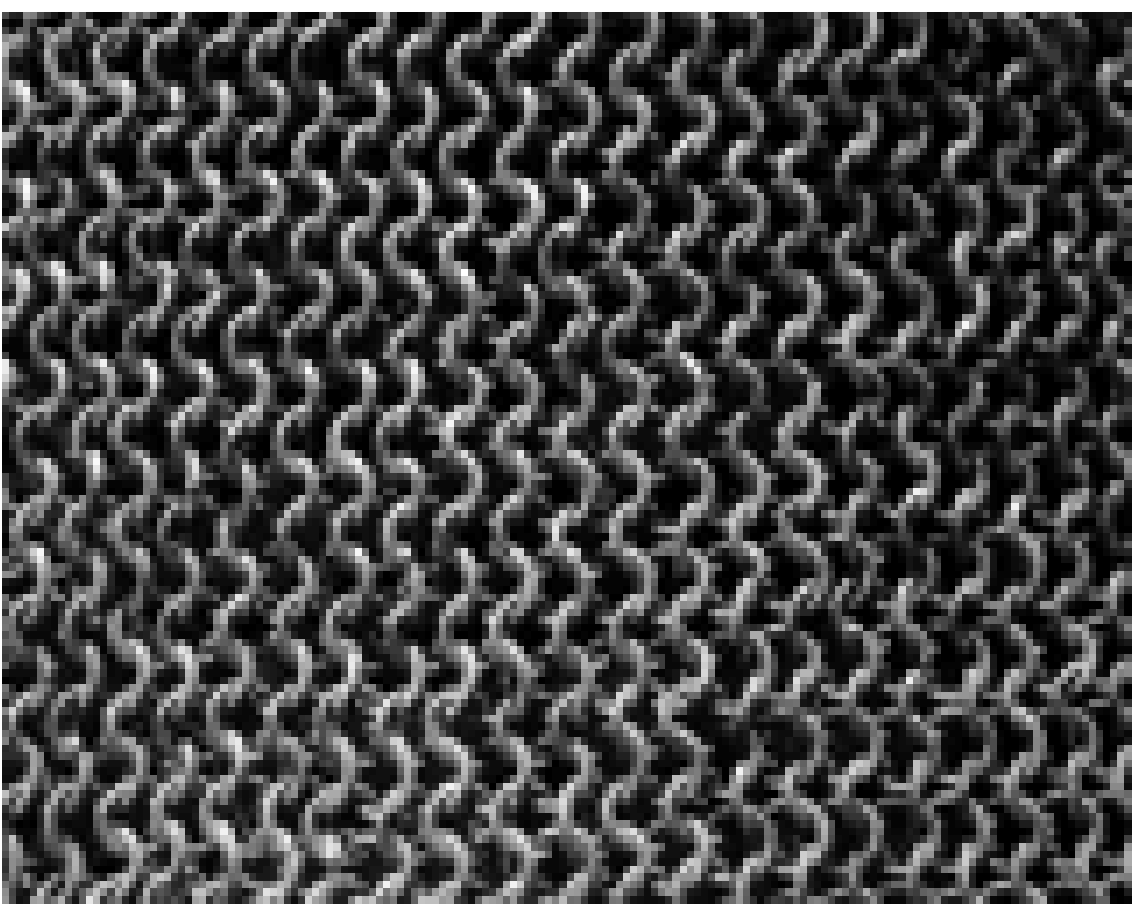} ~~ {\normalsize Chain mail} \quad ~~~{(train = 14101, test = 4779)}}   \\
\midrule
  SMSE & $\bm{0.79}$ &  $1.1$ & $1.1$ & $0.99$ & $0.97$  \\
  MSLL & $\bm{-0.052}$ & $0.036$ & $1.6$ & $0.26$ & $-0.0025$\\
\bottomrule
\end{tabular}
\end{center}
\label{tab: predictions}
\end{table}

We end this Section with a particularly large example, where we use GPatt-10 to perform learning and exact inference on 
the \textit{Pores} pattern, with $383400$ training points, to extrapolate a large missing region with $96600$ 
test points.  The SMSE is $0.077$, and the total runtime was $2800$ seconds.  Images of the successful 
extrapolation are shown in the Appendix.

\subsection{Recovering Complex 3D Kernels from a Video}
\label{sec: recovering}

With a relatively small number of components, GPatt is able to accurately recover a wide range of product kernels.  To test 
GPatt's ability to recover ground truth kernels, we simulate a $50 \times 50 \times 50$ movie of data 
(e.g.\, two spatial input dimensions, one temporal) using a GP with kernel $k = k_1k_2k_3$ (each component kernel 
in this product operates on a different input dimension), where 
$k_1 = k_{\text{SE}}+k_{\text{SE}} \times k_{\text{PER}}$, $k_2 =  k_{\text{MA}} \times k_\text{PER} + k_{\text{MA}} \times k_{\text{PER}}$, 
and $k_3 =  (k_{\text{RQ}}+k_\text{PER})\times k_\text{PER}+ k_\text{SE}$. ($k_{\text{PER}}(\tau) =  \exp[-2\sin^2(\pi\,\tau\,\omega)/\ell^2]$, $\tau = x-x'$).    
We use $5$ consecutive $50\times50$ slices for testing, leaving a large number $N = 112500$ of training points.   In this case,
the big datasize is helpful: the more training instances, the more information to learn the true generating kernels. Moreover, GPatt-20 is 
able to reconstruct these complex out of class kernels in under $10$ minutes.  We compare the learned SMP-20 kernel with 
the true generating kernels in Figure \ref{fig: recoverk}.  In the Appendix, we show true and predicted frames from the movie.

\begin{figure}
\centering%
\includegraphics[width=\linewidth]{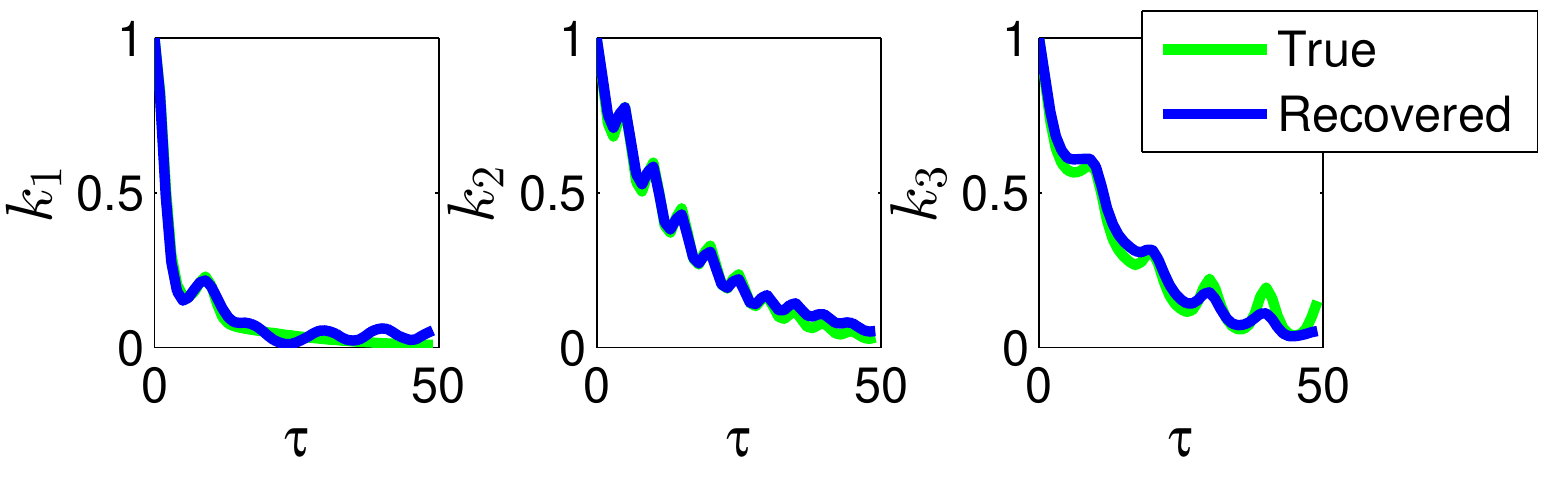}
\caption{Recovering sophisticated product kernels.  A product of three
kernels (shown in green) was used to generate a movie of 112500 3D training points.  From this data, 
GPatt-20 reconstructs these component kernels (the learned SMP-20 kernel is shown in blue).  All 
kernels are a function of $\tau = x-x'$.  For clarity of presentation, each kernel
has been scaled by $k(0)$.}
\label{fig: recoverk}
\end{figure}

\subsection{Wallpaper and Scene Reconstruction}
\label{sec: inpainting}

\begin{figure}
\centering%
\includegraphics[scale=.75]{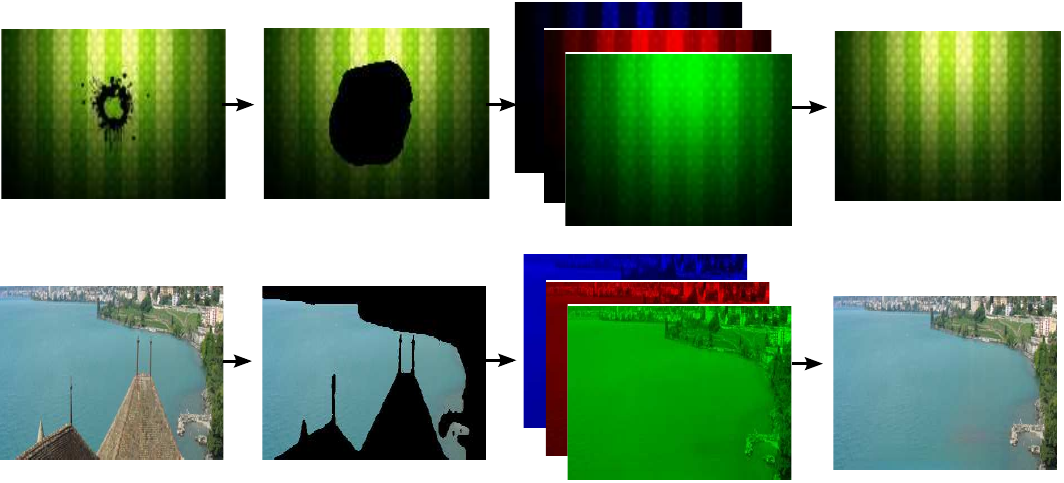}
\caption{Image inpainting with GPatt.  From left to right: A mask is applied
to the original image, GPatt extrapolates the mask 
region in each of the three (red, blue, green) image
channels, and the results are joined to produce the restored image.
Top row: Removing a stain (train: $15047 \times 3$).  Bottom row: Removing a rooftop to restore a natural scene (train: $32269 \times 3$).  The coast
is masked during training and we do not attempt to extrapolate it in testing.}
\label{fig: paintfull}
\end{figure}

Although GPatt is a general purpose regression method, it can also be used for inpainting: image restoration, object removal, etc.

We first consider a wallpaper image stained by a black apple mark, shown in the first row of 
Figure \ref{fig: paintfull}.  
To remove the stain, we apply a mask and then separate the image into its three channels (red, green, and blue). 
This results in $15047$ pixels in each channel for training. In each channel we ran GPatt using SMP-30.  
We then combined the results from each channel to restore the image without any stain, which is particularly 
impressive given the subtleties in the pattern and lighting.

In our next example, we wish to reconstruct a natural scene obscured by a prominent rooftop, shown in the second row
of Figure \ref{fig: paintfull}.  By applying a mask, and following the same procedure as for the stain, this time with $32269$ pixels
in each channel for training, GPatt reconstructs the scene without the rooftop.  This reconstruction captures subtle
details, such as waves in the ocean, even though only one image was used for training.  In fact this 
example has been used with inpainting algorithms which were given access to a repository of thousands of
similar images \citep{hays2008scene}.  The results emphasized that conventional inpainting algorithms and GPatt have profoundly
different objectives, which are sometimes even at cross purposes: inpainting attempts to make the image 
look good to a human (e.g., the example in \citet{hays2008scene} placed boats in the water), while GPatt is a general 
purpose regression algorithm, which simply aims to make accurate predictions at test input locations, from training 
data alone.

\section{Discussion}

Gaussian processes are often used for smoothing and interpolation on small datasets.  However, we believe that Bayesian
nonparametric models are naturally suited to pattern extrapolation on large multidimensional datasets, where extra training 
instances can provide extra opportunities to learn additional structure in data.

The support and inductive biases of a Gaussian process are naturally encoded in a covariance kernel.  A covariance kernel 
must always have some structure to reflect these inductive biases; and that structure can, in principle, be exploited for 
scalable and exact inference, without the need for simplifying approximations.  Such models could play a role in a new era 
of machine learning, where models are expressive and scalable, but also interpretable and manageable, with simple exact learning 
and inference procedures.

We hope to make a small step in this direction with GPatt, a Gaussian process based Bayesian nonparametric framework for 
automatic pattern discovery on large multidimensional datasets, with scalable and exact inference procedures.  Without
human intervention -- no sophisticated initialisation, or hand crafting of kernel features -- GPatt has been used to
accurately and quickly extrapolate large missing regions on a variety of patterns. 

\textbf{Acknowledgements}: We thank Richard E.\ Turner, Ryan Prescott Adams, Zoubin Ghahramani, and Carl Edward Rasmussen
for helpful discussions.

\small 
\bibliography{mbibnew}
\bibliographystyle{icml2014}

\newpage
\onecolumn

\section{Appendix}

\subsection{Introduction}

We provide further detail about the eigendecomposition of kronecker matrices, and the runtime complexity of kronecker matrix vector products.
We also provide spectral images of the learned kernels in the metal tread plate experiment of Section 5.1, larger versions of the images in Table 1, images of the 
extrapolation results on the large pore example, and images of the GPatt reconstruction for several consecutive movie frames.

\subsection{Eigendecomposition of Kronecker Matrices}

Assuming a product kernel,
\begin{equation}
 k(x_i,x_j) = \prod_{p=1}^{P} k^{p}(x_i^p,x_j^p) \,,
\end{equation}
and inputs $x \in \mathcal{X}$ on a multidimensional grid,
$\mathcal{X} = \mathcal{X}_1 \times \dots \times \mathcal{X}_P \subset \mathbb{R}^P$,
then the covariance matrix $K$ decomposes into a Kronecker product of matrices over each
input dimension $K = K^{1} \otimes \dots \otimes K^{P}$ \citep{saatchi11}.
The eigendecomposition of $K$
into $QVQ^{\top}$ similarly decomposes: $Q = Q^{1} \otimes \dots \otimes Q^{P}$ and
$V = V^{1} \otimes \dots \otimes V^{P}$.  Each covariance matrix $K^p$ in the Kronecker
product has entries $K^{p}_{ij} = k^{p}(x_i^p,x_j^p)$ and decomposes as
$K^{p} = Q^{p}V^{p}{Q^{p}}^{\top}$.  Thus the $N \times N$ covariance matrix $K$ can be stored
in $\mathcal{O}(PN^{\frac{2}{P}})$ and decomposed into $QVQ^{\top}$ in $\mathcal{O}(PN^{\frac{3}{P}})$ operations, for $N$ datapoints and
$P$ input dimensions.  \footnote{The total number of datapoints
$N = \prod_p |\mathcal{X}_p|$, where $|\mathcal{X}_p|$ is the cardinality of
$\mathcal{X}_p$.  For clarity of presentation, we assume each $|\mathcal{X}_p|$ has equal cardinality $N^{1/P}$.}

\subsection{Matrix-vector Product for Kronecker Matrices}

We first define a few operators from standard Kronecker literature. Let $\mbf{B}$ be a matrix of size $p \times q$. The $\Reshape(\mbf{B},r,c)$ operator returns a r-by-c matrix ($rc=pq$) whose elements are taken column-wise from $\mbf{B}$. The $\Vect(\cdot)$ operator stacks the matrix columns onto a single vector, $\Vect(\mbf{B})=\Reshape(\mbf{B},pq,1)$, and the $\Vect^{-1}(\cdot)$ operator is defined as $\Vect^{-1}(\Vect(\mbf{\mbf{B}}))=\mbf{B}$. Finally, using the standard Kronecker property $(\mbf{B} \otimes \mbf{C})\Vect(\mbf{X})  = \Vect(\mbf{C}\mbf{X}\mbf{B}^{\top})$, we note that for any $N$ argument vector $\mbf{u} \in \mathbb{R}^N$ we have
{
\begin{align}
\mbf{K}_N\mbf{u} &= \left(\bigotimes_{p=1}^{P} \mbf{K}^{p}_{N^{1/P}}\right)\mbf{u}
=\Vect\left(\mbf{K}^{P}_{N^{1/P}}\mbf{U}\left(\bigotimes_{p=1}^{P-1} \mbf{K}^{p}_{N^{1/P}}\right)^{\top}\right)
,\label{eq:mvprod1}
\end{align}}
\hspace{-3pt}where $\mbf{U} = \Reshape(\mbf{u},N^{1/P},N^{\frac{P-1}{P}})$,
and $\mbf{K}_N$ is an $N \times N$ Kronecker matrix.
With no change to Eq.\ (\ref{eq:mvprod1}) we can introduce the $\Vect^{-1}(\Vect(\cdot))$ operators to get
\begin{align}
\mbf{K}_N\mbf{u} =\Vect\left(\left(~~~\Vect^{-1}\left(\Vect\left(~~~\left(\bigotimes_{p=1}^{P-1} \mbf{K}^{p}_{N^{1/P}}\right)\left(\mbf{K}^P_{N^{1/P}} \mbf{U}\right)^{\top}~~~\right)\right)~~~\right)^{\top}\right). \label{eq:mvprod2}
\end{align}
The inner component of Eq.\ (\ref{eq:mvprod2}) can be written as
\begin{align}
&\Vect\left(\left(\bigotimes_{p=1}^{P-1} \mbf{K}^{p}_{N^{1/P}}\right)\left(\mbf{K}^P_{N^{1/P}} \mbf{U}\right)^{\top}\mbf{I}_{N^{{1}/{P}}}\right) = \mbf{I}_{N^{{1}/{P}}}\otimes\left(\bigotimes_{p=1}^{P-1} \mbf{K}^{p}_{N^{1/P}}\right)\Vect\left(\left(\mbf{K}^P_{N^{1/P}} \mbf{U}\right)^{\top}\right). \label{eq:mvprod3}
\end{align}
Notice that Eq.\ (\ref{eq:mvprod3}) is in the same form as Eq.\ (\ref{eq:mvprod1}) (Kronecker matrix-vector product). By repeating Eqs.\ (\ref{eq:mvprod2}-\ref{eq:mvprod3}) over all $P$ dimensions, and noting that $\left(\bigotimes_{p=1}^{P}\mbf{I}_{N^{{1}/{P}}}\right) \mbf{u} = \mbf{u}$, we see that the original matrix-vector product can be written as
{\small
\begin{align}
\left(\bigotimes_{p=1}^{P} \mbf{K}^{p}_{N^{1/P}}\right)\mbf{u} &= \operatorname{vec}\left(\left[\mbf{K}^{1}_{N^{1/P}},\dots\left[\mbf{K}^{P-1}_{N^{1/P}},\left[\mbf{K}^{P}_{N^{1/P}},\mbf{U}\right]\right]\right]\right) \\
&\; {\buildrel\rm def\over=} \;\Kron\left(\mbf{K}^{1}_{N^{1/P}},\mbf{K}^{2}_{N^{1/P}},\dots,\mbf{K}^{P}_{N^{1/P}}, \mbf{u}\right)
\label{eq:tensorp}
\end{align}
}
\hspace{-3pt}where the bracket notation denotes matrix product, transpose then reshape, i.e.,
\begin{align}
\left[ \mbf{K}^p_{N^{1/P}},\mbf{U} \right] = \Reshape\left(\left(\mbf{K}^p_{N^{1/P}}\mbf{U}\right)^\top,N^{1/P},N^{\frac{P-1}{P}}\right)\,.
\end{align}
Iteratively solving the $\Kron$ operator in Eq.\ (\ref{eq:tensorp}) requires $
(PN^{\frac{P+1}{P}})$, because each of the $P$ bracket operations requires $\mathcal{O}(N^{\frac{P+1}{P}})$. 

\subsection{Inference with Imaginary Observations}

The predictive mean of a Gaussian process at $L$ test points, given $N$ training points, is given by
\begin{align}
 \bm{\mu}_{L} =  \mbf{K}_{LN}\left(\mbf{K}_{N} + \sigma^2 I_N\right)^{-1}\mbf{y}\,, 
\end{align}
where $\mbf{K}_{LN}$ is an $L \times N$ matrix of cross covariances between the test and training points.  We wish to show that when we have
$M$ observations which are not on a grid that the desired predictive mean
\begin{equation}
\label{eqn: muL}
 \bm{\mu}_{L} = \mbf{K}_{LM}\left(\mbf{K}_{M} + \sigma^2 I_M\right)^{-1}\mbf{y}_M =  \mbf{K}_{LN}\left(\mbf{K}_{N} + \mbf{D}_N\right)^{-1}\mbf{y} \,,
\end{equation}
where $\mbf{y} = [\bm{y}_M, \bm{y}_W]^{\top}$ includes imaginary observations $\bm{y}_W$, and $D_N$ is as defined in Section 4.
as
\begin{align}
 {D}_N = \left[ \begin{array}{cc}
{D}_M      &   0 \\
0   &   {\epsilon}^{-1} {I}_W      \end{array} \right],
\end{align}
where we set $D_M = \sigma^2 I_M$.

Starting with the right hand side of Eq.~\eqref{eqn: muL},
\begin{equation}
\label{eqn: bigmuL}
\bm{\mu}_{L}  =  \left[ \begin{array}{c}
\mbf{K}_{LM}   \\
\mbf{K}_{LW}   \end{array} \right]
\left[ \begin{array}{cc}
\mbf{K}_{M} + \mbf{D}_M      &   \mbf{K}_{MW}  \\
\mbf{K}_{MW}^{\top}   &   \mbf{K}_{W}+{\epsilon}^{-1} \mbf{I}_W      \end{array} \right]^{-1}
\left[ \begin{array}{c}
\mbf{y}_M   \\
\mbf{y}_W   \end{array} \right].  \\
\end{equation}

Using the block matrix inversion theorem, we get
{\small
\begin{align}
\label{blk_inv}
&\left[ \begin{array}{cc}
A      &   B  \\
C  &   E     \end{array} \right]^{-1} =
\left[ \begin{array}{cc}
(A-BE^{-1}C)^{-1}     &   -A^{-1}B(I-E^{-1}CA^{-1}B)^{-1}E^{-1}  \\
-E^{-1}C(A-BE^{-1}C)^{-1}   &   (I-E^{-1}CA^{-1}B)^{-1}E^{-1}      \end{array} \right],
\end{align}
}
where $A = \mbf{K}_{M} + \mbf{D}_M $, $B = \mbf{K}_{MW}$, $C=\mbf{K}_{MW}^{\top} $, and $E= \mbf{K}_{W}+{\epsilon}^{-1} \mbf{I}_W$. If we take the limit of $E^{-1}= {\epsilon} (\epsilon\mbf{K}_{W}+\mbf{I}_W)^{-1} \buildrel\epsilon\rightarrow 0\over\longrightarrow \mbf{0}$, and solve for the other components, Eq.~\eqref{eqn: bigmuL}   becomes
\begin{align}
\bm{\mu}_{L}&  =  \left[ \begin{array}{c}
\mbf{K}_{LM}   \\
\mbf{K}_{LW}   \end{array} \right]
\left[ \begin{array}{cc}
\left(\mbf{K}_{M} + \mbf{D}_M\right)^{-1}      &   \mbf{0}  \\
 \mbf{0}   &    \mbf{0}      \end{array} \right]
\left[ \begin{array}{c}
\mbf{y}_M   \\
\mbf{y}_W   \end{array} \right]
= \mbf{K}_{LM}(\mbf{K}_{M} + \mbf{D}_M)^{-1}\mbf{y}_M
\end{align}
which is the exact GP result. In other words, performing inference given observations $\bm{y}$ will give the same result as directly using observations $\bm{y}_M$.
The proof that the predictive covariances remain unchanged proceeds similarly.

\subsection{Spectrum Analysis}
\begin{figure}[!ht]
\centering%
\subfigure[GPatt-30]{\label{fig:spec1}\includegraphics[width=0.3\linewidth]{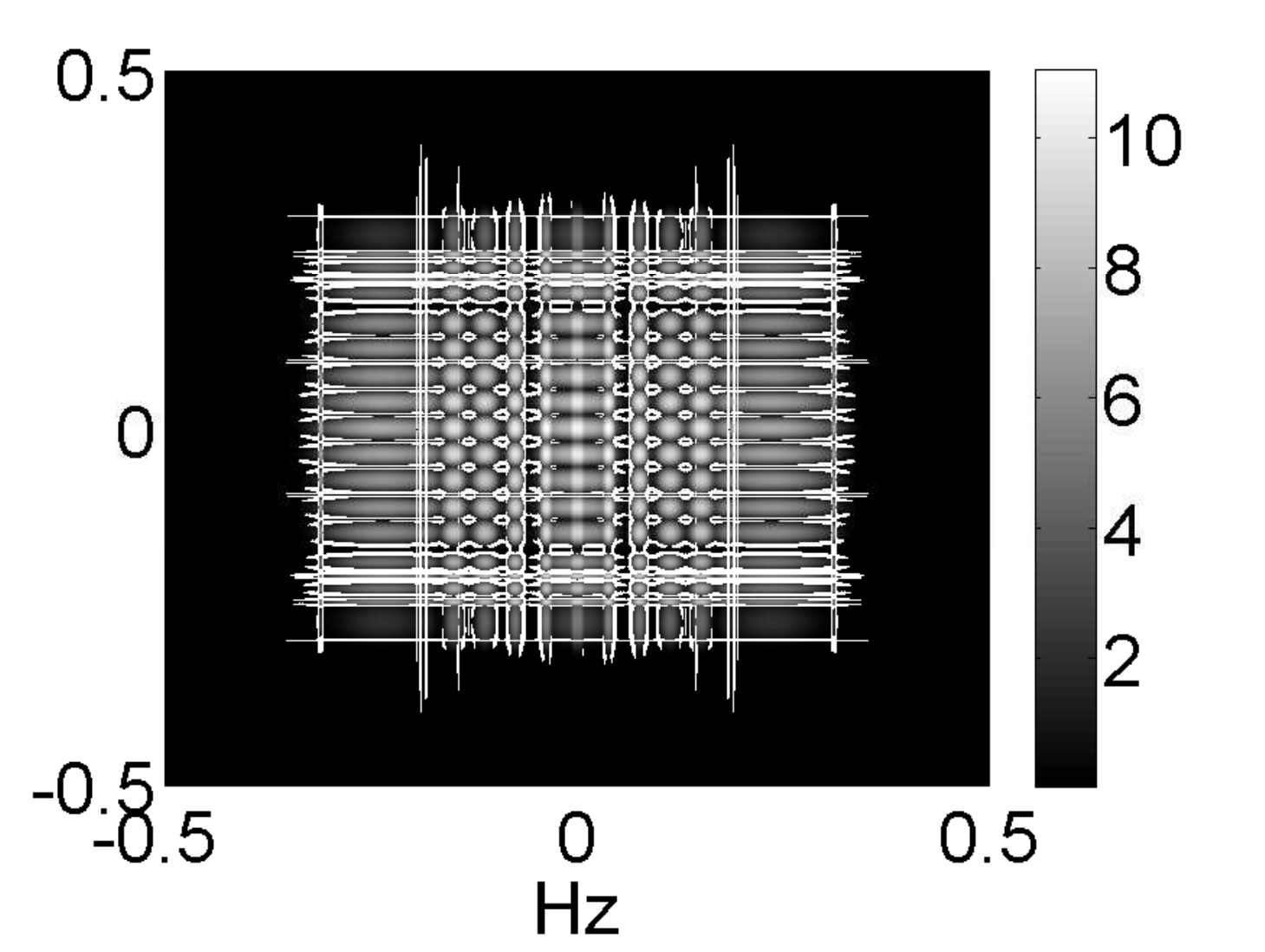}}
\subfigure[FITC]{\label{fig:spec5}\includegraphics[width=0.3\linewidth]{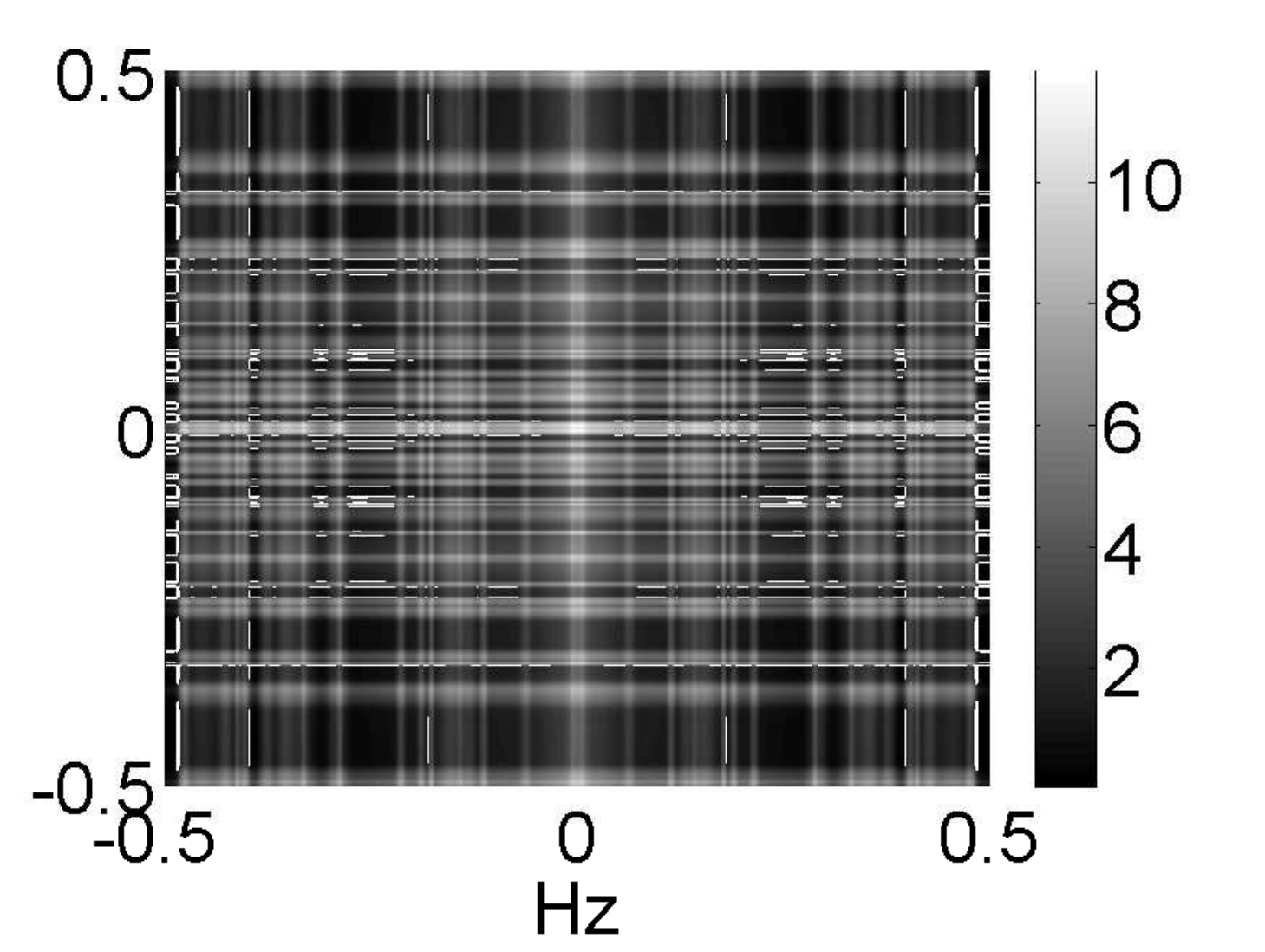}}\\
\subfigure[SE]{\label{fig:spec2}\includegraphics[width=0.3\linewidth]{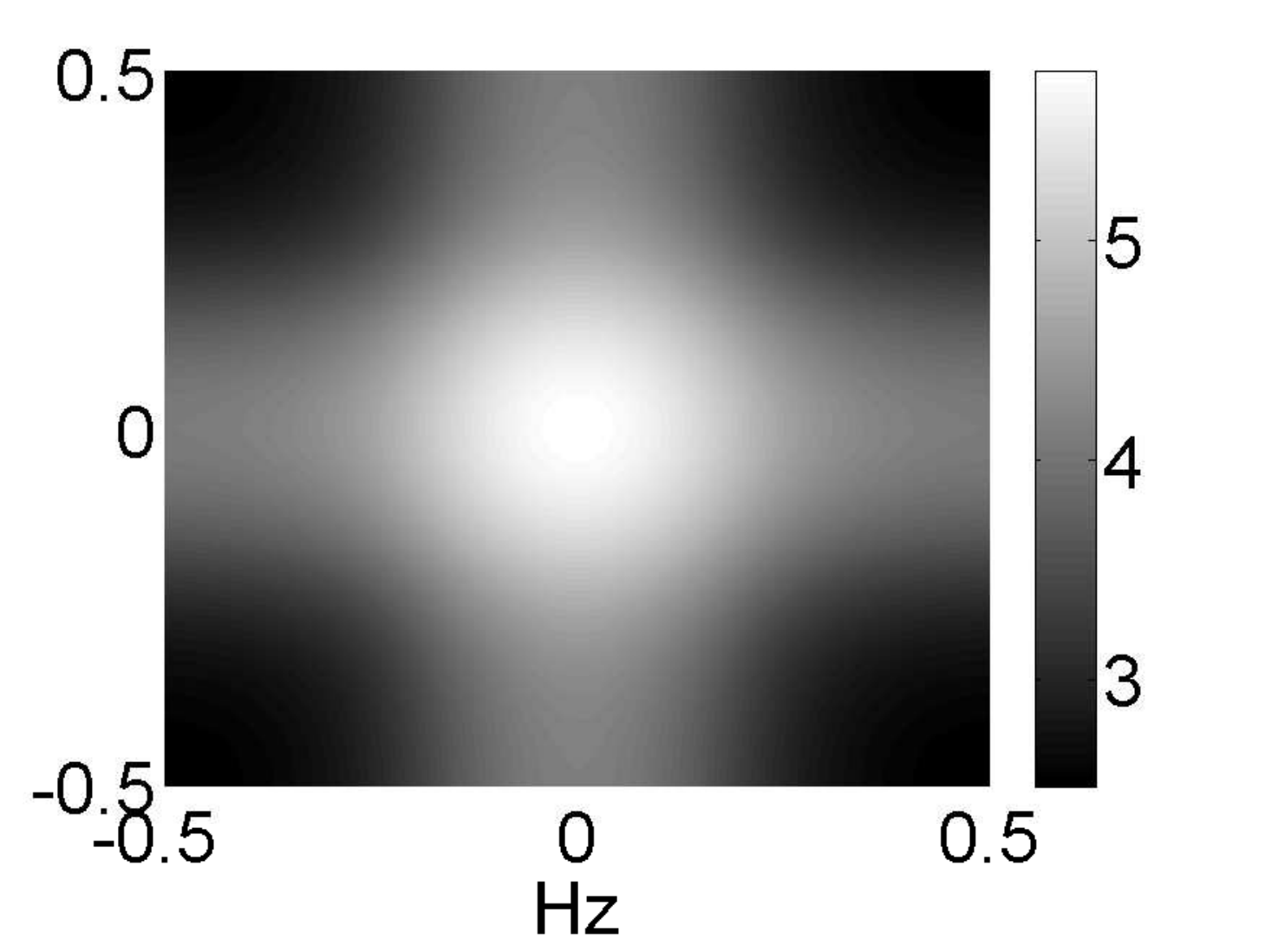}}
\subfigure[RQ]{\label{fig:spec3}\includegraphics[width=0.3\linewidth]{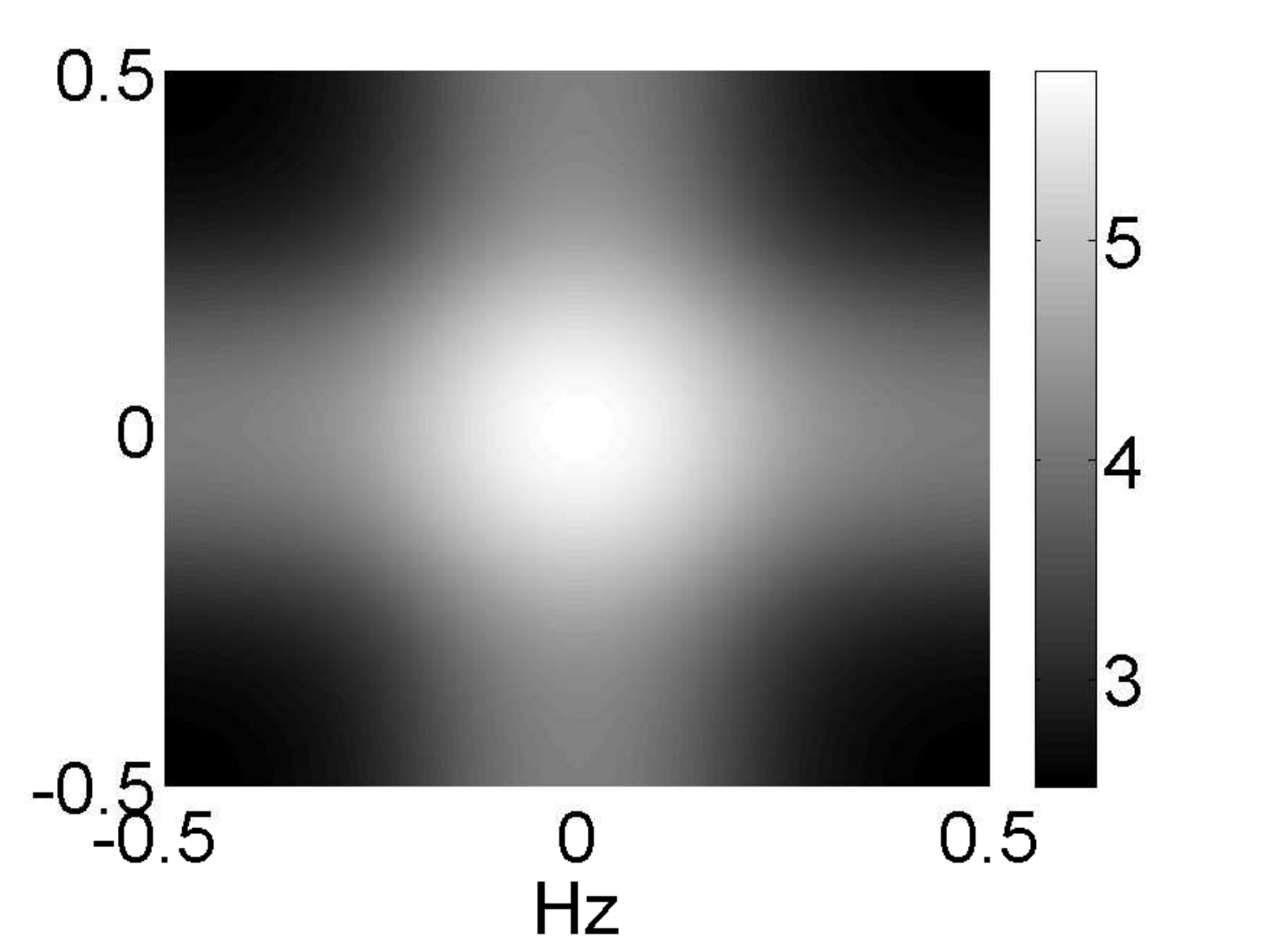}}
\subfigure[MA]{\label{fig:spec4}\includegraphics[width=0.3\linewidth]{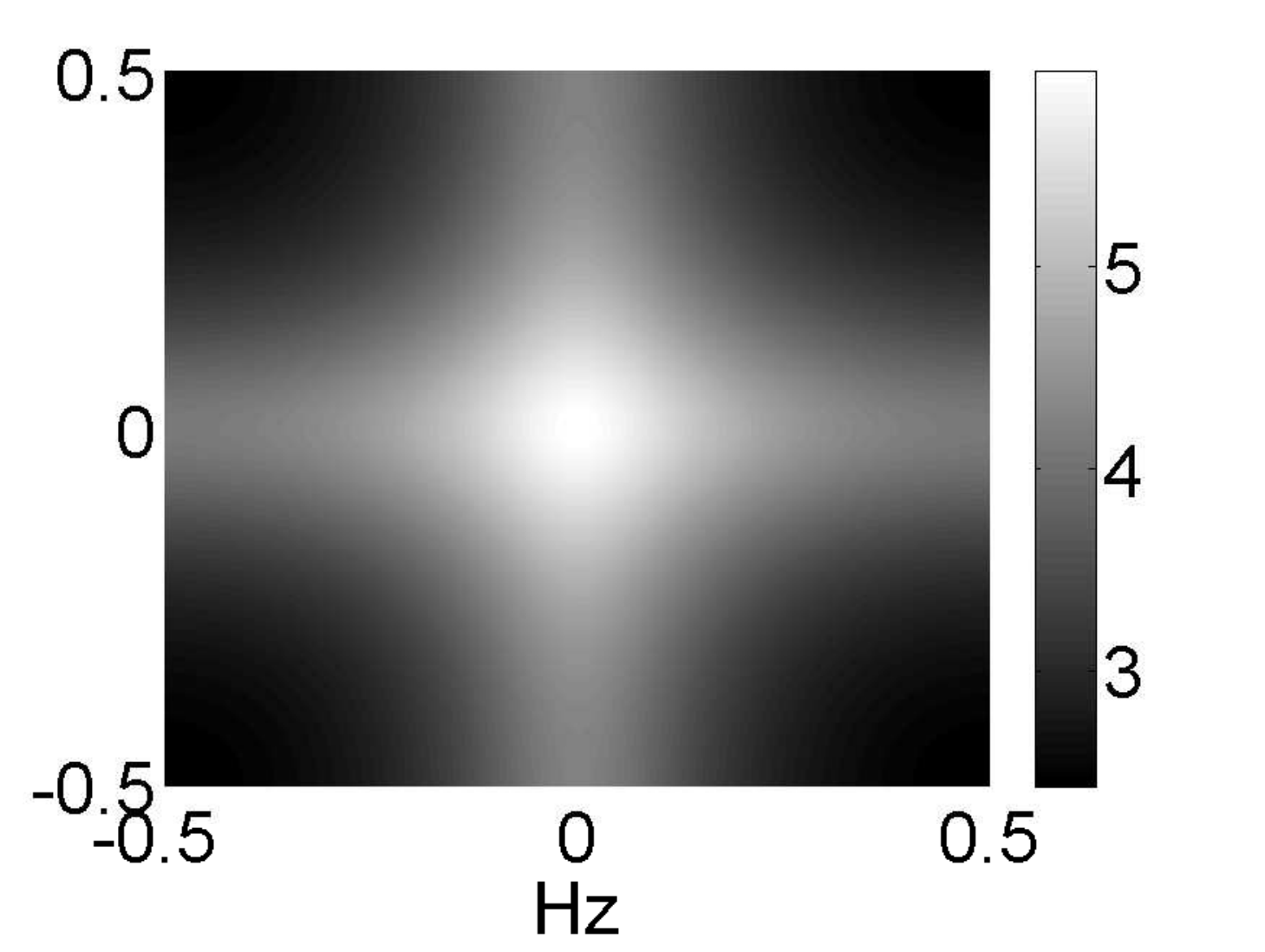}}
\caption{Spectral representation of the learned kernels from Section 5.1. For methods which used the SMP kernel (namely, a) GPatt and b) FITC) we plot the analytical log spectrum using the learned hyperparameters. For c)Squared exponential, d) Rational quadratic, and e) Mat\'ern-3 we plot instead the empirical log spectrum using the Fast
Fourier transform of the kernel.}
\label{fig: spectrum}
\end{figure}

We can gain further insight into the behavior of GPatt by looking at the spectral density learned by the spectral
mixture kernel. Figure \ref{fig: spectrum} shows the log spectrum representations of the learned kernels from Section 5.1.
Smoothers, such as the popular SE, RQ, and MA kernels concentrate their spectral energy around the origin, differing only by their tail support for higher frequencies.  Methods which used the SMP kernel, such as the GPatt and FITC (with an SMP kernel), are able to learn meaningful features in the spectrum space.

\subsection{Images}
\begin{figure}[!ht]
\centering%
\subfigure[Rubber mat]{\label{fig:patt1}\includegraphics[width=0.3\linewidth]{pattern1}}
\subfigure[Tread plate]{\label{fig:patt2}\includegraphics[width=0.3\linewidth]{pattern2}}
\subfigure[Pores]{\label{fig:patt4}\includegraphics[width=0.3\linewidth]{pattern4}}
\subfigure[Wood]{\label{fig:patt5}\includegraphics[width=0.3\linewidth]{pattern5}}
\subfigure[Chain mail]{\label{fig:patt6}\includegraphics[width=0.3\linewidth]{pattern6}}
\subfigure[Cone]{\label{fig:cone}\includegraphics[width=0.3\linewidth]{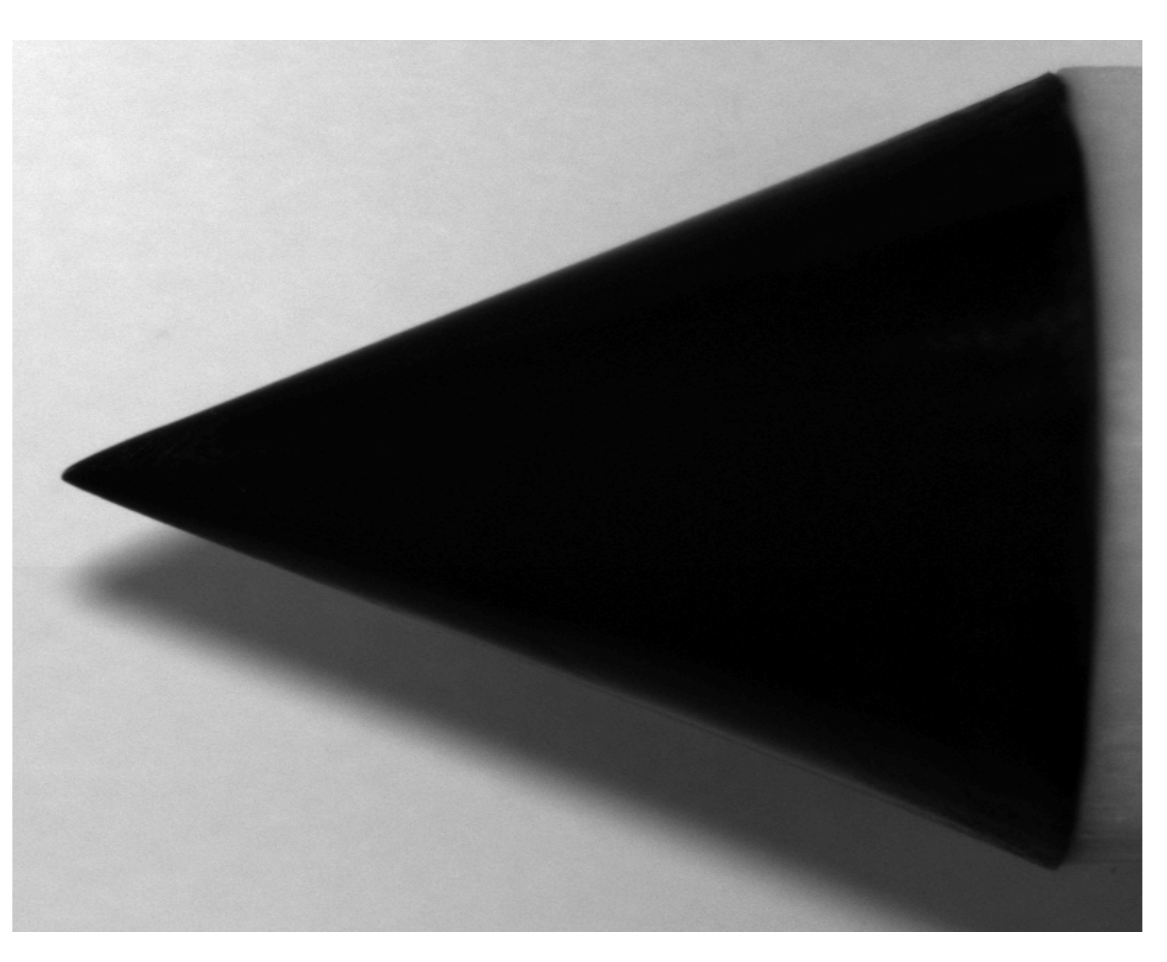}}
\caption{Images used for stress tests in Section 5.2. Figures a) through e) show the textures used in the accuracy comparison of Table 1. Figure e) is the cone image which was used for the runtime analysis shown in Figure 3a.}
\label{fig: pattimages}
\end{figure}

\begin{figure}[!ht]
\centering%
\subfigure[Train]{\label{fig:bigdatatrain}\includegraphics[width=0.49\linewidth]{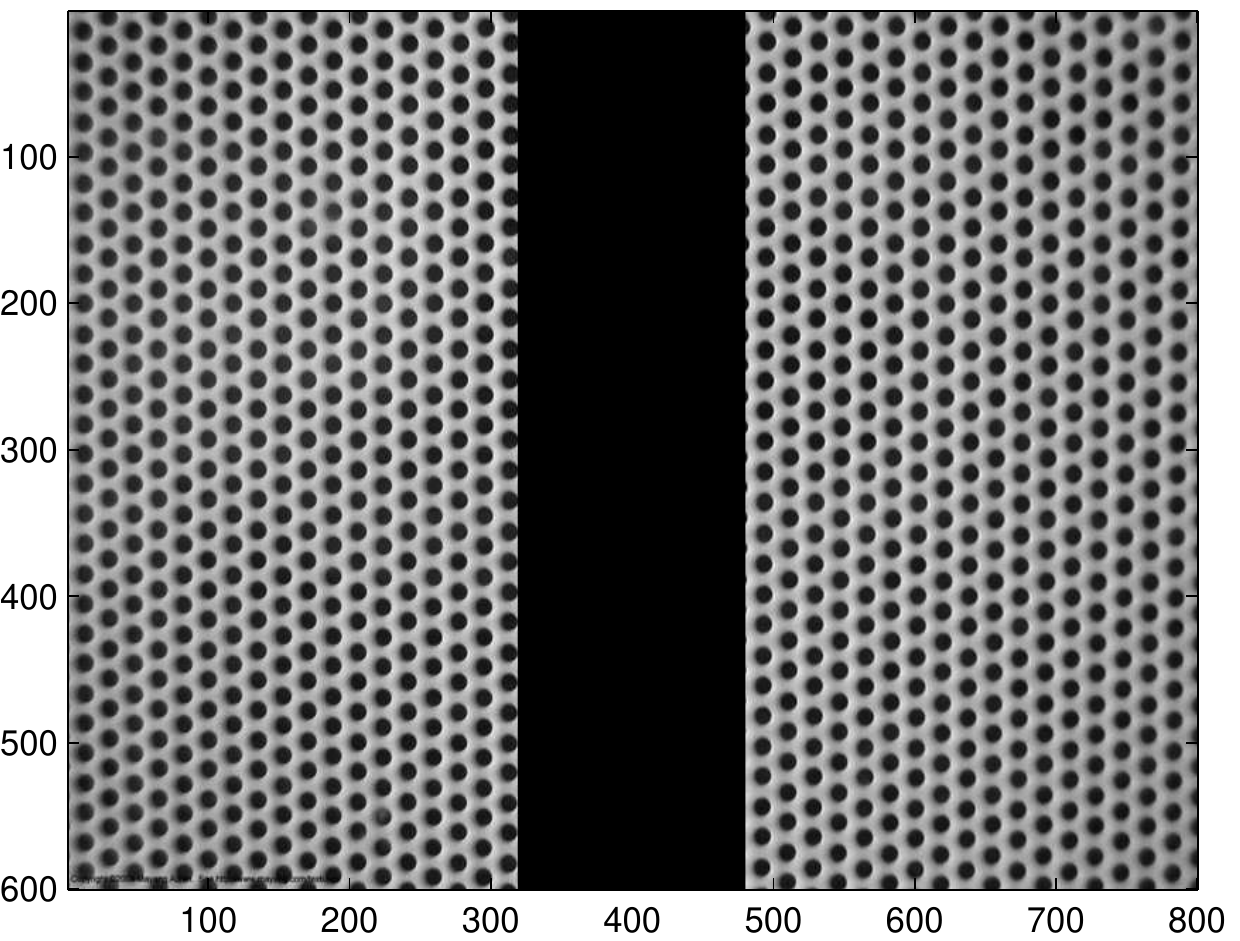}}
\subfigure[Learn]{\label{fig:bigdatalearn}\includegraphics[width=0.49\linewidth]{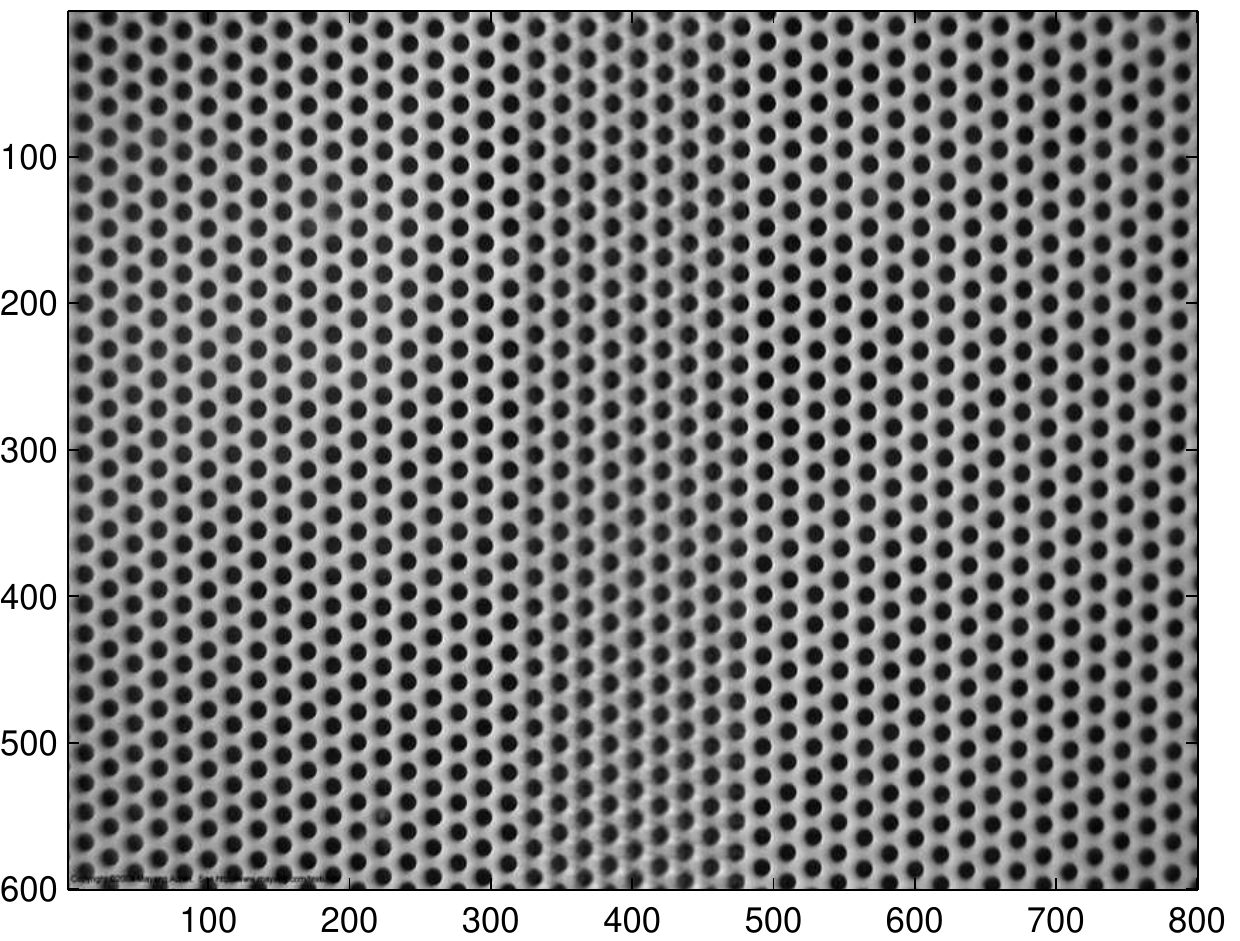}}
\caption{GPatt on a particularly large multidimensional dataset. a) Training region (383400 points), b) GPatt-10 reconstruction of the missing region. }
\label{fig: bigdata}
\end{figure}

\begin{figure}[!ht]
\centering%
\includegraphics[width=0.9\linewidth]{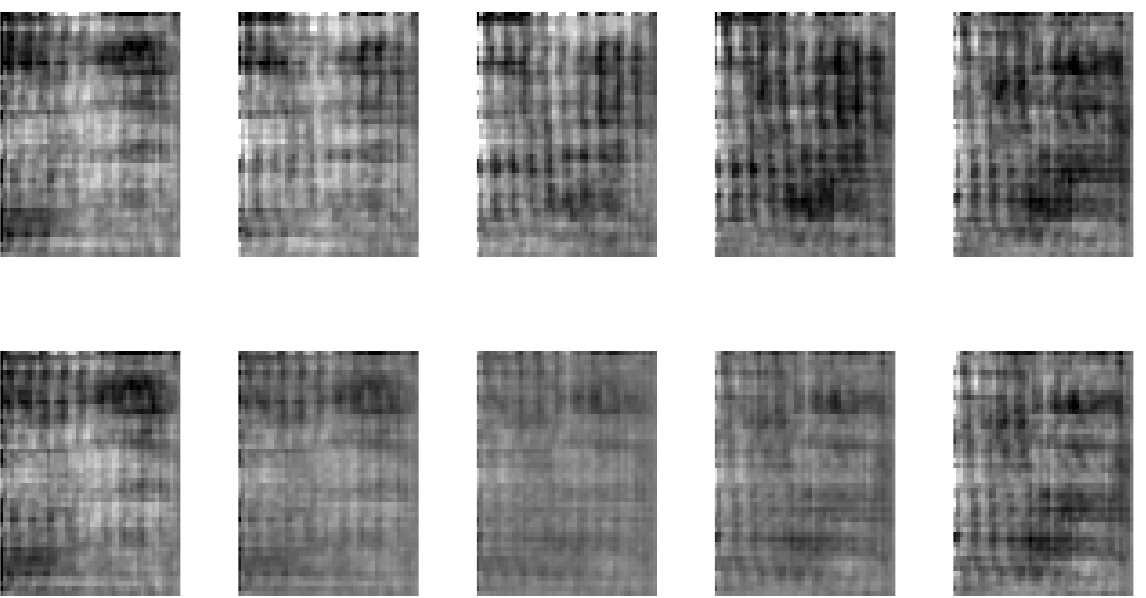}
\caption{Using GPatt to recover 5 consecutive slices from a movie. All slices are missing from training data (e.g., these are not 1 step ahead forecasts).  Top row: true slices take from the middle of the movie. Bottom row: inferred slices using GPatt-20.}
\label{fig: recoverFrames}
\end{figure}

In the rest of the supplementary material, we provide the images and results referenced in the main text. Figure \ref{fig: pattimages} illustrates the images used for the stress tests in Section 5.2. In Figure \ref{fig: bigdata}, we provide the results for the large pore example. Finally, Figure \ref{fig: recoverFrames} shows the true and predicted movie frames discussed in Section 5.3.

\end{document}